\documentclass{article}

\PassOptionsToPackage{numbers, compress}{natbib}
\usepackage[preprint]{nips_2018}

\usepackage[utf8]{inputenc} % allow utf-8 input
\usepackage[T1]{fontenc}    % use 8-bit T1 fonts
\usepackage{hyperref}       % hyperlinks
\usepackage{url}            % simple URL typesetting
\usepackage{booktabs}       % professional-quality tables
\usepackage{amsfonts}       % blackboard math symbols
\usepackage{nicefrac}       % compact symbols for 1/2, etc.
\usepackage{microtype}      % microtypography

\usepackage{amsmath,amsthm,enumitem,caption,multirow,multicol,algorithmic,algorithm,array}
\usepackage{comment}
\usepackage{graphicx}
\usepackage{subcaption}
\usepackage{rotating,booktabs}
\usepackage{afterpage}
\usepackage{tabu}% http://ctan.org/pkg/tabu
\makeatletter
\newcommand{\dashrule}[1][black]{%
	\color{#1}\rule[\dimexpr.5ex-.2pt]{4pt}{.4pt}\xleaders\hbox{\rule{4pt}{0pt}\rule[\dimexpr.5ex-.2pt]{4pt}{.4pt}}\hfill\kern0pt%
}
\usepackage{tikz}
\usetikzlibrary{shapes,arrows, decorations.markings}
\graphicspath{{./figures/}}
\usepackage{changes}
\definechangesauthor[name={Per cusse}, color=red]{per}
\setremarkmarkup{(#2)}

\definecolor{myGreen}{RGB}{0,175,0}

\title{Reinforcement Learning for Solving the Vehicle Routing Problem}

\author{
  Mohammadreza~Nazari\\
  Department of Industrial Engineering\\
  Lehigh University\\
  Bethlehem, PA 18015 \\
  \texttt{mon314@lehigh.edu} \\
   \And
   	Afshin~Oroojlooy \\
   	Department of Industrial Engineering\\
	Lehigh University \\
	Bethlehem, PA 18015 \\
	\texttt{afo214@lehigh.edu} \\
	\And
	Martin Tak\'a\v{c} \\
	Department of Industrial Engineering\\
	Lehigh University \\
	Bethlehem, PA 18015 \\
	\texttt{takac@lehigh.edu} \\
	\And
	Lawrence V.~Snyder \\
	Department of Industrial Engineering\\
	Lehigh University \\
	Bethlehem, PA 18015\\
	\texttt{	lvs2@lehigh.edu} 
}

\begin{document}
% \nipsfinalcopy is no longer used

\maketitle

\begin{abstract}
	We present an end-to-end framework for solving the Vehicle Routing Problem (VRP) using reinforcement learning. In this approach, we train a single model that finds near-optimal solutions for problem instances sampled from a given distribution, only by observing the reward signals and following feasibility rules. Our model represents a parameterized stochastic policy, and by applying a policy gradient algorithm to optimize its parameters, the trained model produces the solution as a sequence of consecutive actions in real time, without the need to re-train for every new problem instance. On capacitated VRP, our approach outperforms classical heuristics and Google's OR-Tools on medium-sized instances in solution quality with comparable computation time (after training). We demonstrate how our approach can handle problems with split delivery and explore the effect of such deliveries on the solution quality. Our proposed framework can be applied to other variants of the VRP such as the stochastic VRP, and has the potential to be applied more generally to combinatorial optimization problems. 
	
\end{abstract}
\section{Introduction}\label{sec:introduction}

The \textit{Vehicle Routing Problem} (VRP) is a combinatorial optimization problem that has been studied in applied mathematics and computer science for decades. VRP is known to be a computationally difficult problem for which many exact and heuristic algorithms have been proposed, but providing fast and reliable solutions is still a challenging task. In the simplest form of the VRP, a single capacitated vehicle is responsible for delivering items to multiple customer nodes; the vehicle must return to the depot to pick up additional items when it runs out. The objective is to optimize a set of routes, all beginning and ending at a given node, called the \textit{depot}, in order to attain the maximum possible reward, which is often the negative of the total vehicle distance or average service time. This problem is computationally difficult to solve to optimality, even with only a few hundred customer nodes \cite{fukasawa2006robust}. For an overview of the VRP, see, for example, \cite{golden2008vehicle,laporte1992vehicle, laporte2000classical,toth2002vehicle}. 

The prospect of new algorithm discovery, without any hand-engineered reasoning, makes neural networks and reinforcement learning a compelling choice that has the potential to be an important milestone on the path toward approaching these problems. In this work, we develop a framework with the capability of solving a wide variety of combinatorial optimization problems using \textit{Reinforcement Learning} (RL) and show how it can be applied to solve the VRP. For this purpose, we consider the Markov Decision Process (MDP) formulation of the problem, in which the optimal solution can be viewed as a sequence of decisions. This allows us to use RL to produce near-optimal solutions by increasing the probability of decoding ``desirable'' sequences. A naive approach is to find a problem-specific solution by considering every instance separately. Obviously, this approach is not practical in terms of either solution quality or runtime since there should be many trajectories sampled from one MDP to be able to produce a near-optimal solution. Moreover, the learned policy does not apply to instances other than the one that was used in the training; with a small perturbation of the problem setting, we need to rebuild the policy from scratch.

Therefore, rather than focusing on training a separate model for every problem instance, we propose a structure that performs well on any problem sampled from a given distribution. This means that if we generate a new VRP instance with the same number of nodes and vehicle capacity, and the same location and demand distributions as the ones that we used during training, then the trained policy will work well, and we can solve the problem right away, without retraining for every new instance. As long as we approximate the generating distribution of the problem, the framework can be applied. One can view the trained model as a black-box heuristic (or a meta-algorithm) which generates a high-quality solution in a reasonable amount of time.

This study is motivated by the recent work by \citet{bello2016neural}. We have generalized their framework to include a wider range of combinatorial optimization problems such as the VRP. \citet{bello2016neural} propose the use of a Pointer Network \cite{vinyals2015pointer} to decode the solution. One major issue that prohibits the direct use of their approach for the VRP is that it assumes the system is static over time. In contrast, in the VRP, the demands change over time in the sense that once a node has been visited its demand becomes, effectively, zero. To overcome this, we propose an alternate approach---which is actually simpler than the Pointer Network approach---that can efficiently handle both the static and dynamic elements of the system. Our model consists of a recurrent neural network (RNN) decoder coupled with an attention mechanism. At each time step, the embeddings of the static elements are the input to the RNN decoder, and the output of the RNN and the dynamic element embeddings are fed into an attention mechanism, which forms a distribution over the feasible destinations that can be chosen at the next decision point.

The proposed framework is appealing since we utilize a self-driven learning procedure that only requires the reward calculation based on the generated outputs; as long as we can observe the reward and verify the feasibility of a generated sequence, we can learn the desired meta-algorithm. For instance, if one does not know how to solve the VRP but can compute the cost of a given solution, then one can provide the signal required for solving the problem using our method. Unlike most classical heuristic methods, it is robust to problem changes, meaning that when the inputs change in any way, it can automatically adapt the solution. Using classical heuristics for VRP, the entire distance matrix must be recalculated and the system must be re-optimized from scratch, which is often impractical, especially if the problem size is large. In contrast, our proposed framework does not require an explicit distance matrix, and only one feed-forward pass of the network will update the routes based on the new data.  

Our numerical experiments indicate that our framework performs significantly better than well-known classical heuristics designed for the VRP, and that it is robust in the sense that its worst results are still relatively close to optimal. Comparing our method with the OR-Tools VRP engine \cite{ortools}, which is ones of the best open-source VRP solvers, we observe a noticeable improvement; in VRP instances with 50 and 100 customers, our method provides shorter tours in roughly $61\%$ of the instances. Another interesting observation that we make in this study is that by allowing multiple vehicles to supply the demand of a single node, our RL-based framework finds policies that outperform the solutions that require single deliveries.  We obtain this appealing property, known as the split delivery, without any hand engineering and no extra cost.

\section{Background}
Before presenting the model, we briefly review some background that is closely related to our work.

\paragraph{Sequence-to-Sequence Models}

\textit{Sequence-to-Sequence} models \cite{sutskever2014sequence,vinyals2015pointer,luong2015effective} are useful in tasks for which a mapping from one sequence to another is required. They have been extensively studied in the field of neural machine translation over the past several years, and there are numerous variants of these models. The general architecture, which is almost the same among different versions, consists of two RNN networks, called the encoder and decoder. An encoder network reads through the input sequence and stores the knowledge in a fixed-size vector representation (or a sequence of vectors); then, a decoder converts the encoded information back to an output sequence. 

In the vanilla Sequence-to-Sequence architecture \cite{sutskever2014sequence}, the source sequence appears only once in the encoder and the entire output sequence is generated based on one vector (i.e., the last hidden state of the encoder RNN). Other extensions, for example \citet{bahdanau2014neural}, illustrate that the source information can be used more wisely to increase the amount of information during the decoding steps. In addition to the encoder and decoder networks, they employ another neural network, namely an \textit{attention mechanism} that \textit{attends} to the entire encoder RNN states. This mechanism allows the decoder to focus on the important locations of the source sequence and use the relevant information during decoding steps for producing ``better'' output sequences. Recently, the concept of attention has been a popular research idea due to its capability to align different objects, e.g., in computer vision \cite{chen2015abc,xiao2015application,xu2015show,hong2016learning} and neural machine translation \cite{bahdanau2014neural,jean2014using, luong2015effective}. In this study, we also employ a special attention structure for policy representation. See Section \ref{sec:vrp:att} for a detailed discussion of the attention mechanism.

\paragraph{Neural Combinatorial Optimization }
Over the last several years, multiple methods have been developed to tackle combinatorial optimization problems by using recent advances in artificial intelligence. The first attempt was proposed by \citet{vinyals2015pointer}, who introduce the concept of a \textit{Pointer Network}, a model originally inspired by sequence-to-sequence models. Because it is invariant to the length of the encoder sequence, the Pointer Network enables the model to apply to combinatorial optimization problems, where the output sequence length is determined by the source sequence. They use the Pointer Network architecture in a supervised fashion to find near-optimal Traveling Salesman Problem (TSP) tours from ground truth optimal (or heuristic) solutions. This dependence on supervision prohibits the Pointer Network from finding better solutions than the ones provided during the training.

Closest to our approach,  \citet{bello2016neural} address this issue by developing a neural combinatorial optimization framework that uses RL to optimize a policy modeled by a Pointer Network. Using several classical combinatorial optimization problems such as TSP and the knapsack problem, they show the effectiveness and generality of their architecture. 

On a related topic, \citet{dai2017learning} solve optimization problems over graphs using a graph embedding structure \cite{dai2016discriminative} and a deep Q-learning (DQN) algorithm \cite{mnih2015human}. Even though VRP can be represented by a graph with weighted nodes and edges, their proposed model does not directly apply since in VRP, a particular node (e.g. the depot) might be visited multiple times. 

Next, we introduce our model, which is a simplified version of the Pointer Network.

\section{The Model}

In this section, we formally define the problem and our proposed framework for a generic combinatorial optimization problem with a given set of inputs $X \doteq \{x^i, i =1,\cdots,M\}$. We allow some of the elements of each input to change between the decoding steps, which is, in fact, the case in many problems such as the VRP. The dynamic elements might be an artifact of the decoding procedure itself, or they can be imposed by the environment. For example, in the VRP, the remaining customer demands change over time as the vehicle visits the customer nodes; or we might consider a variant in which new customers arrive or adjust their demand values over time, independent of the vehicle decisions. Formally, we represent each input $x^i$ by a sequence of tuples $\{x^i_t\doteq(s^i, d^i_t), t=0,1,\cdots\}$, where $s^i$ and $d^i_t$ are the static and dynamic elements of the input, respectively, and can themselves be tuples. One can view $x^i_t$ as a vector of features that describes the state of input $i$ at time $t$. For instance, in the VRP, $x^i_t$ gives a snapshot of the customer $i$, where $s^i$ corresponds to the 2-dimensional coordinate of customer $i$'s location and $d_t^i$ is its demand at time $t$.  We will denote the set of all input states at a fixed time $t$ with $X_t$.

We start from an arbitrary input in $X_0$, where we use the pointer $y_0$ to refer to that input. At every decoding time $t$ ($t=0,1,\cdots$), $y_{t+1}$ points to one of the available inputs $X_t$, which determines the input of the next decoder step; this process continues until a termination condition is satisfied. The termination condition is problem-specific, showing that the generated sequence satisfies the feasibility constraints. For instance, in the VRP that we consider in this work, the terminating condition is that there is no more demand to satisfy. This process will generate a sequence of length $T$, $Y = \{y_t, t = 0, ..., T\}$, possibly with a different sequence length compared to the input length $M$. This is due to the fact that, for example, the vehicle may have to go back to the depot several times to refill. We also use the notation $Y_t$ to denote the decoded sequence up to time $t$, i.e., $Y_t = \{y_0,\cdots,y_t\}$.  We are interested in finding a stochastic policy $\pi$ which generates the sequence $Y$ in a way that minimizes a loss objective while satisfying the problem constraints. The optimal policy $\pi^*$ will generate the optimal solution with probability 1. Our goal is to make $\pi$ as close to $\pi^*$ as possible. Similar to \citet{sutskever2014sequence}, we use the probability chain rule to decompose the probability of generating sequence $Y$, i.e., $P(Y|X_0)$, as follows:
\begin{align}
P(Y|X_0) = \prod_{t=0}^{T} P(y_{t+1} | Y_{t}, X_{t}),\label{eq:prob}
\end{align}
and 
\begin{align}
X_{t+1} = f(y_{t+1}, X_{t})\label{eq:transition}
\end{align}
is a recursive update of the problem representation with the state transition function $f$. Each component in the right-hand side of \eqref{eq:prob} is computed by the attention mechanism, i.e.,
\begin{align}
P(y_{t+1}| Y_{t},{X}_{t}) = \text{softmax}(g(h_t,{X}_{t})),
\end{align}
where $g$ is an affine function that outputs an input-sized vector, and $h_t$ is the state of the RNN decoder that summarizes the information of previously decoded steps $y_0, \cdots,y_t$. We will describe the details of our proposed attention mechanism in Section \ref{sec:vrp:att}.

\textbf{Remark 1}: This model can handle combinatorial optimization problems in both a more classical static setting as well as in dynamically changing ones. In static combinatorial optimization, $X_0$ fully defines the problem that we are trying to solve. For example, in the VRP, $X_0$ includes all customer locations as well as their demands, and the depot location; then, the remaining demands are updated with respect to the vehicle destination and its load. With this consideration, often there exists a well-defined Markovian transition function $f$, as defined in \eqref{eq:transition}, which is sufficient to update the dynamics between decision points. However, our model can also be applied to  problems in which the state transition function is unknown and/or is subject to external noise, since the training does not explicitly make use of the transition function. However, knowing this transition function helps in simulating the environment that the training algorithm interacts with. See Appendix \ref{sec:sto-vrp} for an example of how to apply the model to a stochastic version of the VRP in which random customers with random demands appear over time.

\subsection{Limitations of Pointer Networks}\label{sec:vrp:limit}
Although the framework proposed by \citet{bello2016neural} works well on problems such as the knapsack problem and TSP, it is not applicable to more complicated combinatorial optimization problems in which the system representation varies over time, such as VRP. \citet{bello2016neural} feed a random sequence of inputs to the RNN encoder. Figure \ref{fig:vrp:limit} illustrates with an example why using the RNN in the encoder is restrictive. Suppose that at the first decision step, the policy sends the vehicle to customer 1, and as a result, its demand is satisfied, i.e., $d_0^1\neq d_1^1$. Then in the second decision step, we need to re-calculate the whole network with the new $d_1^1$ information in order to choose the next customer. The dynamic elements complicate the forward pass of the network since there should be encoder/decoder updates when an input changes. The situation is even worse during back-propagation to accumulate the gradients since we need to remember when the dynamic elements changed. In order to resolve this complication, we require the model to be \textit{invariant to the input sequence} so that changing the order of any two inputs does not affect the network. In Section \ref{sec:vrp:proposed_model}, we present a simple network that satisfies this property.

\begin{figure}[tp]
	\centering
	\begin{minipage}{.47\textwidth}
		\centering
	\definecolor{beaublue}{rgb}{0.74, 0.83, 0.9}
\definecolor{almond}{rgb}{0.94, 0.87, 0.8}
\definecolor{staticCol}{rgb}{0.97, 96.0, 1.0}
\definecolor{dynamicCol}{rgb}{0.91, 1.0, 1.0}
\definecolor{coralred}{rgb}{1.0, 0.25, 0.25}

\usetikzlibrary{patterns}

\tikzstyle{rect1} = [rectangle,text width=.4cm, minimum width=.5cm, minimum height=.6cm, text centered, draw=black, fill=white!30, node distance=1.1cm]
\tikzstyle{rect2} = [rectangle,text width=.4cm, minimum width=.5cm, minimum height=1.2cm, text centered, draw=black, fill=white!30, node distance=1.1cm ]
\tikzset{>=latex}
\pgfmathsetmacro{\PHI}{14}
\pgfmathsetmacro{\PHII}{4}
\pgfmathsetmacro{\PSI}{4}
\tikzstyle{vecArrow} = [thick, decoration={markings,mark=at position
   1 with \arrow[semithick]{open triangle 60}},
   double distance=1.4pt, shorten >= 5.5pt,
   preaction = {decorate},
   postaction = {draw,line width=1.4pt, white,shorten >= 4.5pt}]

\tikzset{
  myarrow/.style = {line width=2mm, draw=blue, -triangle 60, fill=blue!40,postaction={draw, line width=4mm, shorten >=6mm, -}}
}

\begin{tikzpicture}[scale=2,node distance=.7cm,scale=1, every node/.style={scale=.8}]

  % slide 1 
  \coordinate (orig) at ( 0.0, 0.0);

  % inputs
  \node (pinp1) [rect1, right of=orig, fill=staticCol] {$s^1$};
  \node (cinp01) [rect1, below of=pinp1, fill=coralred, node distance=.6cm] {$d_1^1$};
  
  \node (pinp2) [rect1, right of=pinp1, fill=staticCol] {$s^2$};
  \node (cinp02) [rect1, below of=pinp2, fill=dynamicCol, node distance=.6cm] {$d_1^2$};
  
  \node (pinp3) [rect1, right of=pinp2, fill=staticCol] {$s^3$};
  \node (cinp03) [rect1, below of=pinp3, fill=dynamicCol, node distance=.6cm] {$d_1^3$};
  
  \node (pinp4) [rect1, right of=pinp3, fill=staticCol] {$s^4$};
  \node (cinp04) [rect1, below of=pinp4, fill=dynamicCol, node distance=.6cm]{$d_1^4$};
  
  \node (pinp5) [rect1, right of=pinp4, fill=staticCol] {$s^5$};
  \node (cinp05) [rect1, below of=pinp5, fill=dynamicCol, node distance=.6cm] {$d_1^5$};

  %encoder
  \node (enc1) [rect1, above of=pinp1, preaction={fill=almond}, pattern=north west lines, pattern color=red] {};
  \node (enc2) [rect1, above of=pinp2, preaction={fill=almond}, pattern=north west lines, pattern color=red] {};
  \node (enc3) [rect1, above of=pinp3, preaction={fill=almond}, pattern=north west lines, pattern color=red] {};
  \node (enc4) [rect1, above of=pinp4, preaction={fill=almond}, pattern=north west lines, pattern color=red] {};
  \node (enc5) [rect1, above of=pinp5, preaction={fill=almond}, pattern=north west lines, pattern color=red] {};
   
  % add horizontal lines
  \draw[black, thick , ->] (enc1) -- (enc2);
  \draw[black, thick , ->] (enc2) -- (enc3);
  \draw[black, thick , ->] (enc3) -- (enc4);
  \draw[black, thick , ->] (enc4) -- (enc5);
  % add vertical lines
  \draw[black, thick , ->] (pinp1) -- (enc1);
  \draw[black, thick , ->] (pinp2) -- (enc2);
  \draw[black, thick , ->] (pinp3) -- (enc3);
  \draw[black, thick , ->] (pinp4) -- (enc4);
  \draw[black, thick , ->] (pinp5) -- (enc5);

  %%%%%%%%%%%%%%%%%%%%%%%%%%%%%%
  % decoder 1
  %%%%%%%%%%%%%%%%%%%%%%%%%%%%%%
  % change of dynamics points
  % add decoder input
  \node (dinp1) [rect2, right of=pinp5, fill=beaublue, yshift=-0.3cm] {$\Rightarrow$};
  \node (dec1) [rect1, right of=enc5, preaction={fill=beaublue}, pattern=north west lines, pattern color=red] {};
  \draw[black, thick , ->] (enc5) -- (dec1);
  \draw[black, thick , ->] (dinp1) -- (dec1);
  
  % add second decoder input
  \node (dpinp2) [rect1, right of=dinp1, yshift=0.30cm,fill=staticCol] {$s^1$};
  \node (dcinp2) [rect1, right of=dinp1,yshift=-0.3cm, fill=dynamicCol ] {$d_0^1$};
  \node (dec2) [rect1, right of=dec1, preaction={fill=beaublue}, pattern=north west lines, pattern color=red] {};
  \draw[black, thick , ->] (dpinp2) -- (dec2);
  
  % draw attention 1
  \coordinate[above of = enc1, yshift=\PHII] (att11) ;
  \coordinate[above of = enc2, yshift=\PHII] (att12) ;
  \coordinate[above of = enc3, yshift=\PHII] (att13) ;
  \coordinate[above of = enc4, yshift=\PHII] (att14) ;
  \coordinate[above of = enc5, yshift=\PHII] (att15) ;
  \coordinate[above of = dec1, yshift=\PHII] (att16) ;
  
  \draw[black, thick , ->,  double ] (att11) -- node[right]{1} (enc1) ;
  \draw[black, thick , ->,shorten >=\PSI pt] (att12) -- (enc2);
  \draw[black, thick , ->,shorten >=\PSI pt] (att13) -- (enc3);
  \draw[black, thick , ->,shorten >=\PSI pt] (att14) -- (enc4);
  \draw[black, thick , ->,shorten >=\PSI pt] (att15) -- (enc5);
  \draw[black, thick ] (dec1) -- (att16);
  \draw[black, thick ] (att16) -- (att11);
  \draw[black, thick , ->] (dec1) -- (dec2);

\end{tikzpicture}
	\caption[Limitation of the Pointer Network]{Limitation of the Pointer Network. After a change in dynamic elements ($d_1^1$ in this example), the whole Pointer Network must be updated to compute the probabilities in the next decision point.}
	\label{fig:vrp:limit}
	\end{minipage}\quad
	\begin{minipage}{.47\textwidth}
	\centering
	\scalebox{0.85 }{
		\hspace{-1.1cm}
	\definecolor{beaublue}{rgb}{0.74, 0.83, 0.9}
\definecolor{staticCol}{rgb}{0.97, 96.0, 1.0}
\definecolor{dynamicCol}{rgb}{0.91, 1.0, 1.0}

\tikzstyle{rect1} = [rectangle,text width=.4cm, minimum width=.5cm, minimum height=.6cm, text centered, draw=black, fill=white!30, node distance=1.1cm]
\tikzstyle{rect2} = [rectangle,text width=.4cm, minimum width=.5cm, minimum height=1cm, text centered, draw=black, fill=white!30, node distance=1.1cm , ]
\tikzstyle{rect3} = [rectangle,text width=3cm, minimum width=4.5cm, minimum height=.5cm, text centered, draw=black, fill=white!30, node distance=5.0cm ]

\tikzset{>=latex}
\pgfmathsetmacro{\PHI}{14}
\pgfmathsetmacro{\PHII}{4}
\pgfmathsetmacro{\PSI}{4}
\tikzstyle{vecArrow} = [thick, decoration={markings,mark=at position
1 with {\arrow[semithick]{open triangle 60}}},
double distance=1.4pt, shorten >= 5.5pt,
preaction = {decorate},
postaction = {draw,line width=1.4pt, white,shorten >= 4.5pt}]
\tikzstyle{myarrows}=[line width=1mm,draw=blue,-triangle 45,postaction={draw, line width=3mm, shorten >=4mm, -}]

% draw two rectangles sticking together
\tikzset{
 data/.style={
 draw,
 rectangle split,
 rectangle split parts=2,
 text centered,
 minimum width=.6 cm,
 node distance=1.1cm
 }
}

\begin{tikzpicture}[scale=2,node distance=.7cm,scale=1, every node/.style={scale=.8}]
  \coordinate (orig) at ( -0.0, -0.0);
  % inputs
  \node (pinp1) [rect1, right of=orig, fill=staticCol] {$\bar{s}^1$};
  \node (dinp1) [rect1, below of=pinp1, fill=dynamicCol, node distance=.6cm] {$\bar{d}^1_1$};
  
  \node (pinp2) [rect1, right of=pinp1, fill=staticCol] {$\bar{s}^2$};
  \node (dinp2) [rect1, below of=pinp2, fill=dynamicCol, node distance=.6cm] {$\bar{d}^2_1$};
  
  \node (pinp3) [rect1, right of=pinp2, fill=staticCol] {$\bar{s}^3$};
  \node (dinp3) [rect1, below of=pinp3, fill=dynamicCol, node distance=.6cm] {$\bar{d}^3$};
  
  \node (pinp4) [rect1, right of=pinp3, fill=staticCol] {$\bar{s}^4$};
  \node (dinp4) [rect1, below of=pinp4, fill=dynamicCol, node distance=.6cm] {$\bar{d}^4_1$};
  
  \node (pinp5) [rect1, right of=pinp4, fill=staticCol] {$\bar{s}^5$};
  \node (dinp5) [rect1, below of=pinp5, fill=dynamicCol, node distance=.6cm] {$\bar{d}^5_1$};

  % add decoder nodes
  \node (dec1) [rect1, right of=pinp5,  fill=beaublue] {};
  \node (dec2) [rect1, right of=dec1, fill=white,draw=none] {};
  
  % attention
  \node (att) [rect1, above of=pinp2, minimum width=8cm, minimum height =2.3cm , yshift = 1.cm, xshift =.4cm, dashed,draw=red]{};
  \node (attText) [right of=att, xshift= 1.4cm,yshift=.5cm,text width=1cm, align=right]{\textit{Attention layer}};
  
  \node (a) [rectangle, above of=dinp5, minimum width =.5cm, minimum height =.6cm,draw=black, fill=white!30, yshift=1.5 cm]{};
  \node (aa) [right of=a]{$a_t$};

  % add horizontal lines
  \draw[black, thick , ->,dotted] (pinp1.north) -- (a.south);
  \draw[black, thick , ->,dotted] (pinp2.north) -- (a.south);
  \draw[black, thick , ->,dotted] (pinp3.north) -- (a.south);
  \draw[black, thick , ->,dotted] (pinp4.north) -- (a.south);
  \draw[black, thick , ->,dotted] (pinp5.north) -- (a.south);
  \draw[black, thick , ->,dotted] (dec1.north) -- (a.south);
  
  \coordinate[above of = a, yshift=\PHII] (out) ;
  % demands
  \node (ddinp1) [rect1, below of=dec1, fill=staticCol] {$\bar{s}^1$};

  % add vertical lines
  \draw[black, thick , ->] (ddinp1) -- (dec1);

  \draw[black, thick , ->] (dec1) -- (dec2);

  % add context
  \coordinate[above of = pinp3, yshift=0.7cm, xshift =-.2cm] (context) ;
    \node (c11) [rect1, above of=context, fill=staticCol] {};
  \node (c12) [rect1, below of=c11, fill=dynamicCol, node distance=.6cm] {};
    \node (ct) [right of=c11,yshift=-.35cm]{$c_t$};
    
  % add horizontal lines
  \draw[black, thick , ->,dashed] (pinp1.north) -- (c12.south);
  \draw[black, thick , ->,dashed] (pinp2.north) -- (c12.south);
  \draw[black, thick , ->,dashed] (pinp3.north) -- (c12.south);
  \draw[black, thick , ->,dashed] (pinp4.north) -- (c12.south);
  \draw[black, thick , ->,dashed] (pinp5.north) -- (c12.south);
  \draw[black, thick , ->,dashed] (a.west) -- (c12.east);
  
 % probabilities
   \node (prob) [rectangle, above of=pinp1, minimum width =.5cm, minimum height =.6cm,draw=black, fill=white!30, yshift=1.6 cm,xshift=-.2cm]{};
 \node (proba) [left of=prob,xshift=-.4cm]{$P(y_{t+1}|.)$};

 % add horizontal lines
 \draw[black, thick , ->] (pinp1.north) -- (prob.south);
 \draw[black, thick , ->] (pinp2.north) -- (prob.south);
 \draw[black, thick , ->] (pinp3.north) -- (prob.south);
 \draw[black, thick , ->] (pinp4.north) -- (prob.south);
 \draw[black, thick , ->] (pinp5.north) -- (prob.south);
 \draw[black, thick , ->] ([yshift=3.2pt]c12.west) -- (prob.east);

  \coordinate[above of=prob,yshift=1 cm] (out2)  ;
 \draw[black, thick , ->] (prob.north) -- (out2);

  % inputs
\node (ppinp1) [rect1, below of=pinp1, fill=staticCol,yshift=-.5 cm] {$s^1$};
\node (cinp01) [rect1, below of=ppinp1, fill=dynamicCol, node distance=.6cm] {$d_t^1$};

\node (ppinp2) [rect1, right of=ppinp1, fill=staticCol] {$s^2$};
\node (cinp02) [rect1, below of=ppinp2, fill=dynamicCol, node distance=.6cm] {$d_t^2$};

\node (ppinp3) [rect1, right of=ppinp2, fill=staticCol] {$s^3$};
\node (cinp03) [rect1, below of=ppinp3, fill=dynamicCol, node distance=.6cm] {$d_t^3$};

\node (ppinp4) [rect1, right of=ppinp3, fill=staticCol] {$s^4$};
\node (cinp04) [rect1, below of=ppinp4, fill=dynamicCol, node distance=.6cm]{$d_t^4$};

\node (ppinp5) [rect1, right of=ppinp4, fill=staticCol] {$s^5$};
\node (cinp05) [rect1, below of=ppinp5, fill=dynamicCol, node distance=.6cm] {$d_t^5$};

\draw[black, thick , ->] (ppinp1) -- (dinp1);
\draw[black, thick , ->] (ppinp2) -- (dinp2);
\draw[black, thick , ->] (ppinp3) -- (dinp3);
\draw[black, thick , ->] (ppinp4) -- (dinp4);
\draw[black, thick , ->] (ppinp5) -- (dinp5);

% embedding
\node (attbox) [rectangle, minimum width = 7.5cm , left of=orig, draw=red, minimum height=1.5cm,dashed,yshift=-.3cm, xshift=3.0cm] {};
\node (atttext) [left of=attbox,xshift=-2cm] {$Embedding$};
\end{tikzpicture}
	}
	\caption{Our proposed model. The embedding layer maps the inputs to a high-dimensional vector space. On the right, an RNN decoder stores the information of the decoded sequence. Then, the RNN hidden state and embedded input produce a probability distribution over the next input using the attention mechanism.}
	\label{fig:our}
	\end{minipage}
\end{figure}

\subsection{The Proposed Neural Network Model}	\label{sec:vrp:proposed_model}

We argue that the RNN encoder adds extra complication to the encoder but is actually not necessary, and the approach can be made much more general by omitting it. RNNs are necessary only when the inputs convey sequential information; e.g., in text translation the combination of words and their relative position must be captured in order for the translation to be accurate. But the question here is, \textit{why do we need to have them in the encoder for combinatorial optimization problems when there is no meaningful order in the input set?} As an example, in the VRP, the inputs are the set of unordered customer locations with their respective demands, and their order is not meaningful; any random permutation contains the same information as the original inputs. Therefore, in our model, we simply leave out the encoder RNN and directly use the embedded inputs instead of the RNN hidden states. By this modification, many of the computational complications disappear, without decreasing the model's efficiency. In Appendix \ref{sec:comp-with-pointer}, we provide an experiment to verify this claim.

As illustrated in Figure \ref{fig:our}, our model is composed of two main components. The first is a set of embeddings that maps the inputs into a $D$-dimensional vector space. We might have multiple embeddings corresponding to different elements of the input, but they are shared among the inputs. The second component of our model is a decoder that points to an input at every decoding step. As is common in the literature \cite{bahdanau2014neural,sutskever2014sequence,cho2014learning}, we use RNN to model the decoder network. Notice that we feed the static elements as the inputs to the decoder network. The dynamic element can also be an input to the decoder, but our experiments on the VRP do not suggest any improvement by doing so, so dynamic elements are used only in the attention layer, described next.

\subsection{Attention Mechanism}\label{sec:vrp:att}
An attention mechanism is a differentiable structure for addressing different parts of the input. Figure \ref{fig:our} illustrates the attention mechanism employed in our method. At decoder step $i$, we utilize a context-based attention mechanism with glimpse, similar to \citet{vinyals2015order}, which extracts the relevant information from the inputs using a variable-length alignment vector $a_t$. In words, $a_t$ specifies how much every input data point might be relevant in the next decoding step $t$.

Let $\bar{x}^i_t = (\bar{s}^i,\bar{d}^i_t)$ be the embedded input $i$, and $h_t\in \mathbb{R}^D$ be the memory state of the RNN cell at decoding step $t$. The alignment vector $a_t$ is then computed as
\begin{equation} 
a_t = a_t(\bar{x}^i_t ,  h_t) = \text{softmax} \left(u_t\right), \label{eq:vrp:align}\quad \text{ where }u_t^i = v^T_a\tanh\left( W_a [\bar{x}^i_t;h_t ]\right).
\end{equation}
Here ``;'' means the concatenation of two vectors. We compute the conditional probabilities by combining the context vector $c_t$, computed as 
\begin{equation}
c_t = \sum_{i=1}^{M} a_t^i \bar{x}_t^i, \label{eq:vrp:ct}
\end{equation}
with the embedded inputs, and then normalizing the values with the softmax function, as follows:
\begin{align}
P(y_{t+1}| Y_{t},{X}_{t}) = 
\text{softmax}(\tilde{u}_t^i) , \quad \text{ where }\tilde{u}_t^i = v^T_c\tanh\left(W_c [\bar{x}_t^i;c_t]\right) \label{eq:vrp:probct}.
\end{align}
In \eqref{eq:vrp:align}--\eqref{eq:vrp:probct}, $v_a$, $v_c$, $W_a$ and $W_c$ are trainable variables.

\textbf{Remark 2}: \textit{Model Symmetry}: \citet{vinyals2015order} discuss an extension of sequence-to-sequence models where they empirically demonstrate that in tasks with no obvious input sequence, such as sorting, the order in which the inputs are fed into the network matter. A similar concern arises when using Pointer Networks for combinatorial optimization problems. However, the model proposed in this paper does not suffer from such a complication since the embeddings and the attention mechanism are invariant to the input order.

\subsection{Training Method}
To train the network, we use well-known policy gradient approaches. To use these methods, we parameterize the stochastic policy $\pi$ with parameters $\theta$. Policy gradient methods use an estimate of the gradient of the expected return with respect to the policy parameters to iteratively improve the policy. In principle, the policy gradient algorithm contains two networks: (\textit{i}) an actor network that predicts a probability distribution over the next action at any given decision step, and (\textit{ii}) a critic network that estimates the reward for any problem instance from a given state. Our training methods are quite standard, and due to space limitation we leave the details to the Appendix.

\section{Computational Experiment}\label{sec:vrp-experiment}

Many variants of the VRP have been extensively studied in the operations research literature. (See, for example, the reviews by \citet{laporte1992vehicle,laporte2000classical}, or the book by \citet{toth2002vehicle} for different variants of the problem.) In this section, we consider a specific capacitated version of the problem in which one vehicle with a limited capacity is responsible for delivering items to many geographically distributed customers with finite demands. When the vehicle's load runs out, it returns to the depot to refill. We will denote the vehicle's remaining load at time $t$ as $l_t$. The objective is to minimize the total route length while satisfying all of the customer demands. This problem is often called the capacitated VRP (CVRP) to distinguish it from other variants, but we will refer to it simply as the VRP.

We assume that the node locations and demands are randomly 
generated from a fixed distribution. Specifically, the customers and depot locations are randomly generated in the unit square $[0,1] \times [0,1]$. For simplicity of exposition, we assume that the demand of each node is a discrete number in $\{1,..,9\}$, chosen uniformly at random. We note, however, that the demand values can be generated from any distribution, including continuous ones.

We assume that the vehicle is located at the depot at time 0, so the first input to the decoder is an embedding of the depot location. At each decoding step, the vehicle chooses from among the customer nodes or the depot to visit in the next step. After visiting customer node $i$, the demands and vehicle load are updated as follows: % according to \eqref{eq:vrp:dem}-\eqref{eq:vrp:load},
\begin{align}
d_{t+1}^i = \max(0,d_{t}^i-l_t),\quad  d_{t+1}^k = d_{t}^k\;\text{   for }k\neq i, \text{and }\quad l_{t+1} = \max(0,l_t-d_{t}^i) \label{eq:vrp:dem-load} 
\end{align}
which is an explicit definition of the state transition function \eqref{eq:transition} for the VRP.

In this experiment, we have employed two different decoders:
\textit{(i)} greedy, in which at every decoding step, the node (either customer or depot) with the highest probability is selected as the next destination, and \textit{(ii)} beam search (BS), which keeps track of the most probable paths and then chooses the one with the minimum tour length \cite{neubig2017neural}. Our results indicate that by applying the beam search algorithm, the quality of the solutions can be improved with only a slight increase in computation time. 

For faster training and generating feasible solutions, we have used a \textit{masking scheme} which sets the log-probabilities of infeasible solutions to $-\infty$ or forces a solution if a particular condition is satisfied. In the VRP, we use the following masking procedures: \textit{(i)} nodes with zero demand are not allowed to be visited; \textit{(ii)} all customer nodes will be masked if the vehicle's remaining load is exactly 0; and \textit{(iii)} the customers whose demands are greater than the current vehicle load are masked. Notice that under this masking scheme, the vehicle must satisfy all of a customer's demands when visiting it. We note, however, that if the situation being modeled does allow split deliveries, one can relax \textit{(iii)}. Indeed, the relaxed masking allows split deliveries, so the solution can allocate the demands of a given customer into multiple routes. This property is, in fact, an appealing behavior that is present in many real-world applications but is often neglected in classical VRP algorithms. In all the experiments of the next section, we do not allow to split demands. Further investigation and illustrations of this property is included in Appendix \ref{sec:split-demand}--\ref{sec:sample-vrp-solution}.

\subsection{Results}

In this section, we compare the solutions found using our framework with those obtained from the \textit{Clarke-Wright savings heuristic} (CW), the \textit{Sweep heuristic} (SW), and Google's optimization tools (OR-Tools). We run our test on multiple problem sizes with different vehicle capacities; for example, VRP10 consists of 10 customer. The results are based on 1000 instances, sampled for each problem size.

\begin{figure*}[htbp]
	\centering
	\begin{subfigure}{.43\columnwidth}
		\includegraphics[width=\columnwidth]{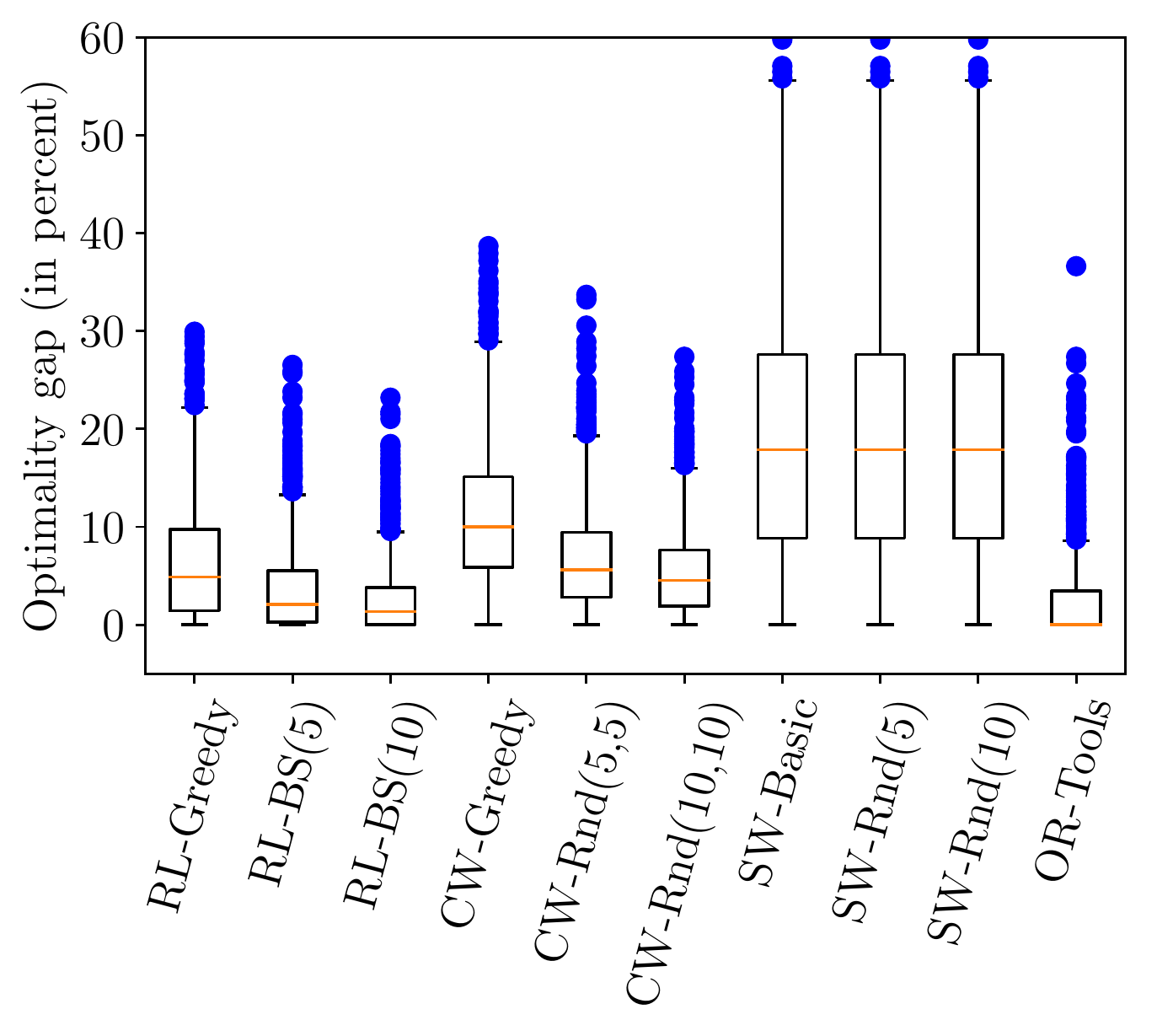}
		\caption{Comparison for VRP10}
		\label{vrp:gap:compa}
	\end{subfigure}
	\begin{subfigure}{.43\columnwidth}
		\includegraphics[width=\columnwidth]{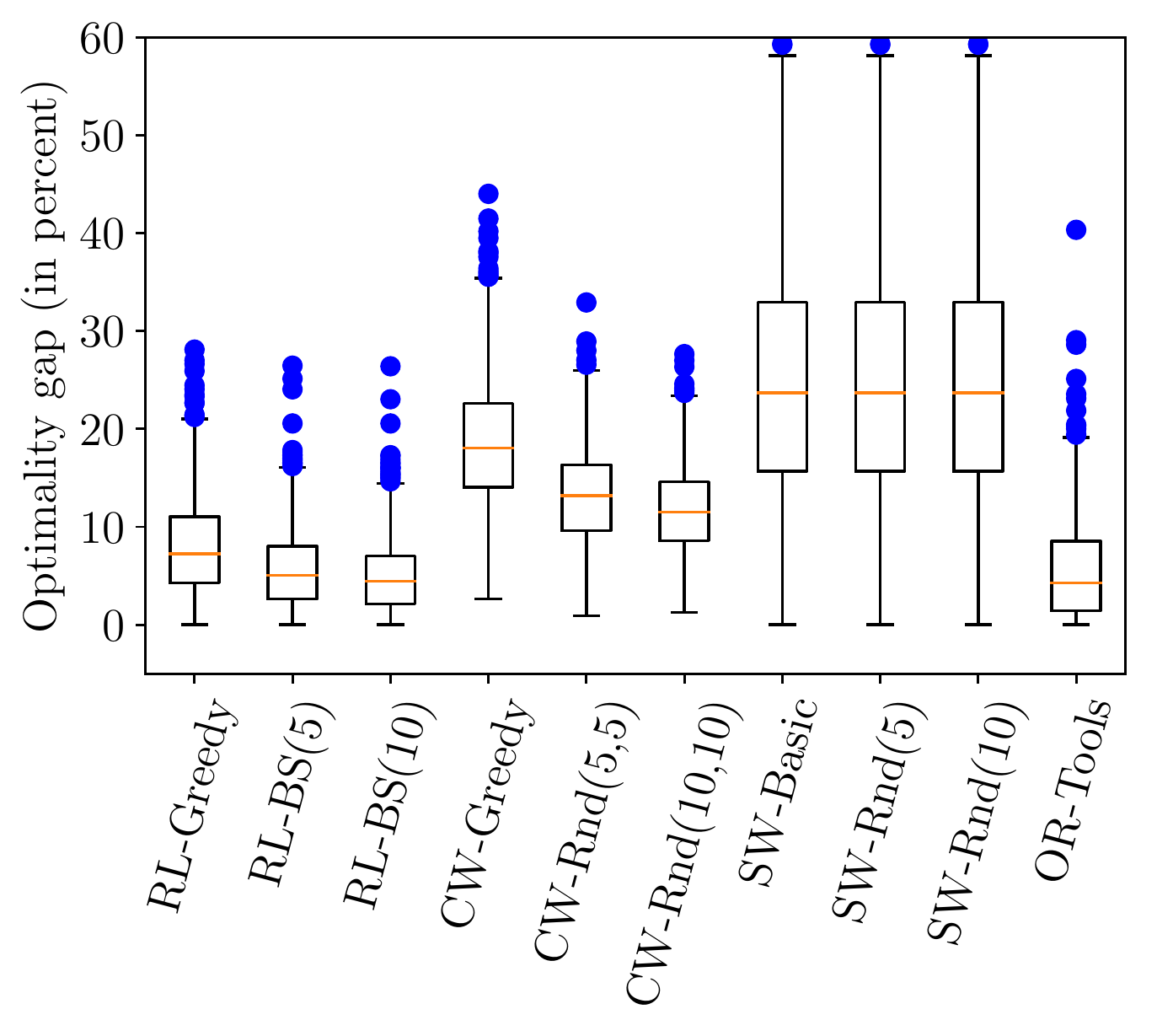}
		\caption{Comparison for VRP20}
		\label{vrp:gap:compb}
	\end{subfigure}
	\begin{subfigure}{.49\columnwidth}
		\includegraphics[width=1.05\columnwidth]{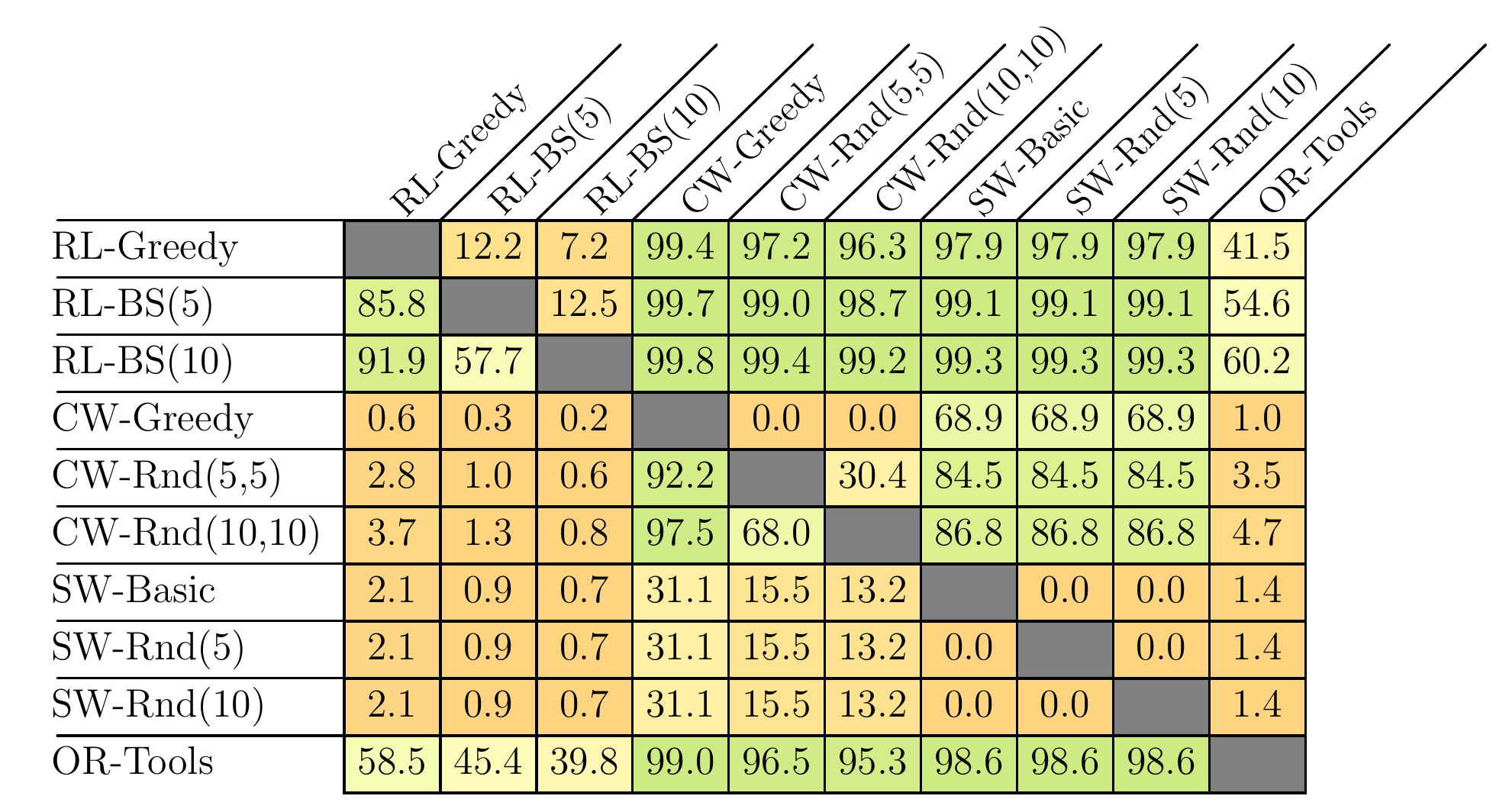}
		\caption{Comparison for VRP50}
		\label{vrp:frac:compc}
	\end{subfigure}
	\begin{subfigure}{.49\columnwidth}
		\includegraphics[width=1.05\columnwidth]{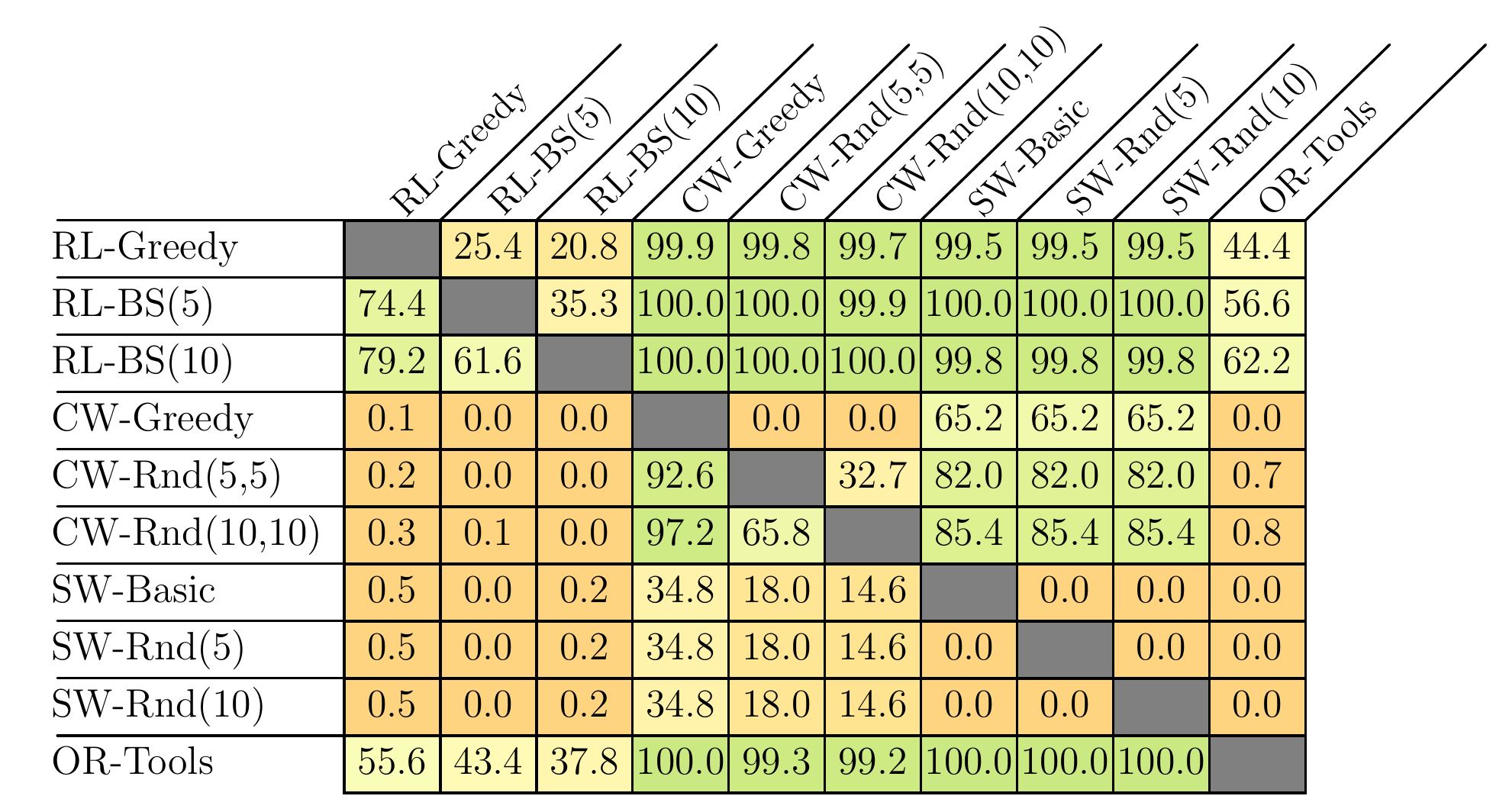}
		\caption{Comparison for VRP100}
		\label{vrp:frac:compd}
	\end{subfigure}

	\caption{Parts \ref{vrp:gap:compa} and \ref{vrp:gap:compb} show the optimality gap (in percent) using different algorithms/solvers for VRP10 and VRP20. Parts \ref{vrp:frac:compc} and \ref{vrp:frac:compd} give the proportion of the samples for which the algorithms in the rows outperform those in the columns; for example, RL-BS(5) is superior to RL-greedy in 85.8\% of the VRP50 instances. }
	\label{fig:vrp:cvrp-comp}
\end{figure*}

Figure \ref{fig:vrp:cvrp-comp} shows the distribution of total tour lengths generated by our method, using greedy and BS decoders, with the number inside the parentheses indicating the beam-width parameter. In the experiments, we label our method with the ``RL'' prefix. In addition, we also implemented a randomized version of both heuristic algorithms to improve the solution quality; for Clarke-Wright, the numbers inside the parentheses are the randomization depth and randomization iterations parameters; and for Sweep, it is the number of random initial angles for grouping the nodes. Finally, we use Google's OR-Tools \cite{ortools}, which is a more competitive baseline. See Appendix \ref{sec:vrp:heuristic} for a detailed discussion on the baselines.

For small problems of VRP10 and VRP20, it is possible to find the optimal solution, which we do by solving a mixed integer formulation of the VRP \cite{toth2002vehicle}. Figures \ref{vrp:gap:compa} and \ref{vrp:gap:compb} measure how far the solutions are far from optimality. The optimality gap is defined as the distance from the optimal objective value normalized by the latter. We observe that using a beam width of 10 is the best-performing method; roughly 95\% of the instances are at most 10\% away from optimality for VRP10 and 13\% for VRP20. Even the outliers are within 20--25\% of optimality, suggesting that our RL-BS methods are robust in comparison to the other baseline approaches. %We also observe that our method can effectively use split delivery to obtain solutions that are 5--10\% shorter than the optimal tours in the classical VRP formulation.} \mycomment{Wait. Now I am realizing this is a problem. You can't compare the results of an algorithm that allows split deliveries with an algorithm that does not. It would be like claiming that your MIP solver is better than other MIP solvers because it only solves the LP relaxation, so the optimal objective function values are better! Somehow I never realized this was the comparison you are making, and I'm sorry about that. I can think of three approaches we can take at this point: (1) Impose single-delivery constraints (i.e., don't allow split deliveries) and rerun all the experiments. This doesn't seem possible by tomorrow. (2) Keep things the way they are, hope the referees don't notice or don't object too much, and plan to fix it in the revision. (3) In the solutions returned by your algorithm, how many demand nodes actually have their demands split among multiple vehicles? My guess is the answer is ``not many.'' (A similar fact is known to be true when we relax single-sourcing constraints in facility location problems.) If that's the case, you can note that although some of the difference between your algorithm and the benchmarks is due to the relaxation of the single-delivery constraint, this effect is very small (give numbers, if possible). If I were a referee, I would still insist that you compare apples to apples, that is, that you enforce single deliveries when comparing to the benchmarks; but I wouldn't be too angry at you because you are acknowledging that you are comparing apples to oranges and are doing your best to mitigate the discrepancy.}
	
Since obtaining the optimal objective values for VRP50 and VRP100 is not computationally affordable, in Figures \ref{vrp:frac:compd} and \ref{vrp:frac:compd}, we compare the algorithms in terms of their winning rate. Each table gives the percentage of the instances in which the algorithms in the rows outperform those in the columns. In other words, the cell corresponding to (A,B) shows the percentage of the samples in which algorithm A provides shorter tours than B. We observe that the classical heuristics are outperformed by the other approaches in almost 100\% of the samples. Moreover, RL-greedy is comparable with OR-Tools, but incorporating beam search into our framework increases the winning rate of our approach to above 60\%.

\begin{figure}[ht]
	\centering
	\includegraphics[width=.48\columnwidth]{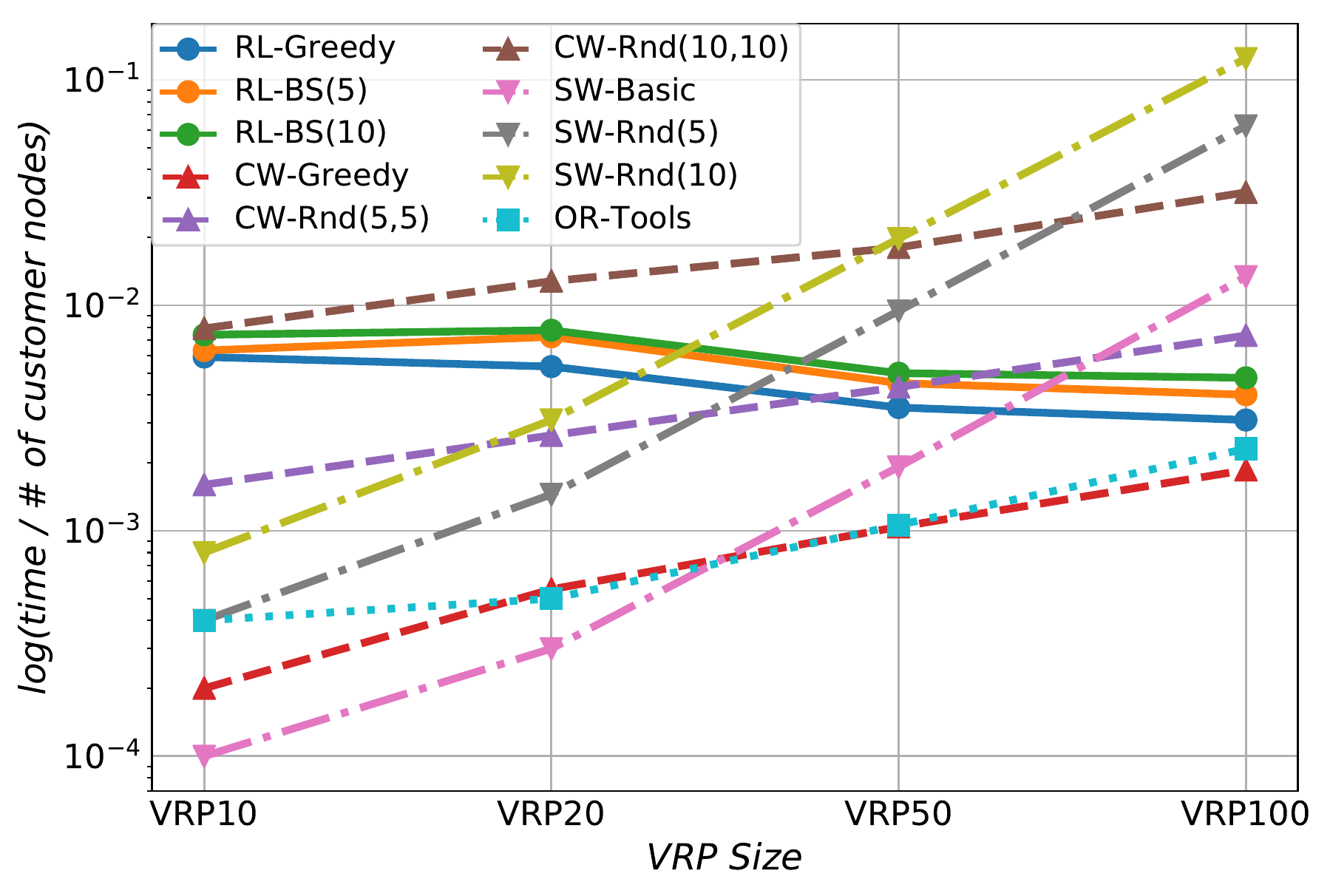}
	\caption{Log of ratio of solution time to the number of customer nodes using different algorithms.}
	\label{fig:vrp:times}
\end{figure}

Figure \ref{fig:vrp:times} shows the log of the ratio of solution times to the number of customer nodes. We observe that this ratio stays almost the same for RL with different decoders. In contrast, the log of the run time for the Clarke-Wright and Sweep heuristics increases faster than linearly with the number of nodes. This observation is one motivation for applying our framework to more general combinatorial problems, since it suggests that our method scales well. Even though the greedy Clark-Wright and basic Sweep heuristics are fast for small instances, they do not provide competitive solutions. Moreover, for larger problems, our framework is faster than the randomized heuristics. We also include the solution times for OR-Tools in the graph, but we should note that OR-Tools is implemented in C++, which makes exact time comparisons impossible since the other baselines were implemented in Python.

\subsection{Extension to Other VRPs}
The proposed framework can be extended easily to problems with multiple depots; one only needs to construct the corresponding state transition function and masking procedure. It is also possible to include various side constraints: soft constraints can be applied by penalizing the rewards, or hard constraints such as time windows can be enforced through a masking scheme. However, designing such a scheme might be a challenging task, possibly harder than solving the optimization problem itself. Another interesting extension is for VRPs with multiple vehicles. In the simplest case in which the vehicles travel independently, one must only design a shared masking scheme to avoid the vehicles pointing to the same customer nodes. Incorporating competition or collaboration among the vehicles is also an interesting line of research that relates to multi-agent RL (MARL) \cite{bucsoniu2010multi}.

This framework can also be applied to real-time services including on-demand deliveries and taxis. In Appendix \ref{sec:sto-vrp}, we design an experiment to illustrate the performance of the algorithm on a VRP where both customer locations and their demands are subject to change. Our results indicate superior performance than the baselines.

\section{Discussion and Conclusion}
We expect that the proposed architecture has significant potential to be used in real-world problems with further improvements. Noting that the proposed algorithm is not limited to VRP, it will be an important topic of future research to apply it to other combinatorial optimization problems such as bin-packing, job-shop, and flow-shop.

This method is quite appealing since the only requirement is a verifier to find feasible solutions and also a reward signal to demonstrate how well the policy is working. Once the trained model is available, it can be used many times, without needing to re-train for the new problems as long as they are generated from the training distribution. Unlike many classical heuristics, our proposed method scales well with increasing problem size, and has a superior performance with competitive solution-time. It doesn't require a distance matrix calculation which might be computationally cumbersome, especially in dynamically changing VRPs. We also illustrate the performance of the algorithm on a much more complicated stochastic version of the VRP.

\section*{Acknowledgment}
This work is supported by U.S. National Science Foundation, under award number NSF:CCF:1618717, NSF:CMMI:1663256 and NSF:CCF:1740796.

\nocite{}

\bibliographystyle{plainnat}
\bibliography{research}

\begin{thebibliography}{38}
\providecommand{\natexlab}[1]{#1}
\providecommand{\url}[1]{\texttt{#1}}
\expandafter\ifx\csname urlstyle\endcsname\relax
  \providecommand{\doi}[1]{doi: #1}\else
  \providecommand{\doi}{doi: \begingroup \urlstyle{rm}\Url}\fi

\bibitem[Applegate et~al.(2006)Applegate, Bixby, Chvatal, and
  Cook]{applegate2006traveling}
David~L Applegate, Robert~E Bixby, Vasek Chvatal, and William~J Cook.
\newblock \emph{The traveling salesman problem: a computational study}.
\newblock Princeton university press, 2006.

\bibitem[Archetti and Speranza(2008)]{archetti2008split}
Claudia Archetti and Maria~Grazia Speranza.
\newblock The split delivery vehicle routing problem: a survey.
\newblock In \emph{The vehicle routing problem: Latest advances and new
  challenges}, pages 103--122. Springer, 2008.

\bibitem[Bahdanau et~al.(2015)Bahdanau, Cho, and Bengio]{bahdanau2014neural}
Dzmitry Bahdanau, Kyunghyun Cho, and Yoshua Bengio.
\newblock Neural machine translation by jointly learning to align and
  translate.
\newblock \emph{In International Conference on Learning Representations}, 2015.

\bibitem[Bello et~al.(2016)Bello, Pham, Le, Norouzi, and
  Bengio]{bello2016neural}
Irwan Bello, Hieu Pham, Quoc~V Le, Mohammad Norouzi, and Samy Bengio.
\newblock Neural combinatorial optimization with reinforcement learning.
\newblock \emph{arXiv preprint arXiv:1611.09940}, 2016.

\bibitem[Bu{\c{s}}oniu et~al.(2010)Bu{\c{s}}oniu, Babu{\v{s}}ka, and
  De~Schutter]{bucsoniu2010multi}
Lucian Bu{\c{s}}oniu, Robert Babu{\v{s}}ka, and Bart De~Schutter.
\newblock Multi-agent reinforcement learning: An overview.
\newblock In \emph{Innovations in multi-agent systems and applications-1},
  pages 183--221. Springer, 2010.

\bibitem[Chen et~al.(2015)Chen, Wang, Chen, Gao, Xu, and Nevatia]{chen2015abc}
Kan Chen, Jiang Wang, Liang-Chieh Chen, Haoyuan Gao, Wei Xu, and Ram Nevatia.
\newblock Abc-cnn: An attention based convolutional neural network for visual
  question answering.
\newblock \emph{arXiv preprint arXiv:1511.05960}, 2015.

\bibitem[Cho et~al.(2014)Cho, Van~Merri{\"e}nboer, Gulcehre, Bahdanau,
  Bougares, Schwenk, and Bengio]{cho2014learning}
Kyunghyun Cho, Bart Van~Merri{\"e}nboer, Caglar Gulcehre, Dzmitry Bahdanau,
  Fethi Bougares, Holger Schwenk, and Yoshua Bengio.
\newblock Learning phrase representations using rnn encoder-decoder for
  statistical machine translation.
\newblock \emph{Conference on Empirical Methods in Natural Language
  Processing}, 2014.

\bibitem[Christofides(1976)]{christofides1976worst}
Nicos Christofides.
\newblock Worst-case analysis of a new heuristic for the travelling salesman
  problem.
\newblock Technical report, Carnegie-Mellon Univ Pittsburgh Pa Management
  Sciences Research Group, 1976.

\bibitem[Clarke and Wright(1964)]{clarke1964scheduling}
Geoff Clarke and John~W Wright.
\newblock Scheduling of vehicles from a central depot to a number of delivery
  points.
\newblock \emph{Operations research}, 12\penalty0 (4):\penalty0 568--581, 1964.

\bibitem[Dai et~al.(2016)Dai, Dai, and Song]{dai2016discriminative}
Hanjun Dai, Bo~Dai, and Le~Song.
\newblock Discriminative embeddings of latent variable models for structured
  data.
\newblock In \emph{International Conference on Machine Learning}, pages
  2702--2711, 2016.

\bibitem[Dai et~al.(2017)Dai, Khalil, Zhang, Dilkina, and
  Song]{dai2017learning}
Hanjun Dai, Elias~B Khalil, Yuyu Zhang, Bistra Dilkina, and Le~Song.
\newblock Learning combinatorial optimization algorithms over graphs.
\newblock \emph{Advances in Neural Information Processing Systems}, 2017.

\bibitem[Fukasawa et~al.(2006)Fukasawa, Longo, Lysgaard, de~Arag{\~a}o, Reis,
  Uchoa, and Werneck]{fukasawa2006robust}
Ricardo Fukasawa, Humberto Longo, Jens Lysgaard, Marcus~Poggi de~Arag{\~a}o,
  Marcelo Reis, Eduardo Uchoa, and Renato~F Werneck.
\newblock Robust branch-and-cut-and-price for the capacitated vehicle routing
  problem.
\newblock \emph{Mathematical programming}, 106\penalty0 (3):\penalty0 491--511,
  2006.

\bibitem[Glorot and Bengio(2010)]{glorot2010understanding}
Xavier Glorot and Yoshua Bengio.
\newblock Understanding the difficulty of training deep feedforward neural
  networks.
\newblock In \emph{Proceedings of the Thirteenth International Conference on
  Artificial Intelligence and Statistics}, pages 249--256, 2010.

\bibitem[Glover and Laguna(2013)]{glover2013tabu}
Fred Glover and Manuel Laguna.
\newblock Tabu search*.
\newblock In \emph{Handbook of combinatorial optimization}, pages 3261--3362.
  Springer, 2013.

\bibitem[Golden et~al.(2008)Golden, Raghavan, and Wasil]{golden2008vehicle}
Bruce~L Golden, Subramanian Raghavan, and Edward~A Wasil.
\newblock \emph{The Vehicle Routing Problem: Latest Advances and New
  Challenges}, volume~43.
\newblock Springer Science \& Business Media, 2008.

\bibitem[Google(2018)]{ortools}
Inc. Google.
\newblock Google's optimization tools (or-tools), 2018.
\newblock URL \url{https://github.com/google/or-tools}.

\bibitem[Gurobi~Optimization(2016)]{gurobi}
Inc. Gurobi~Optimization.
\newblock Gurobi optimizer reference manual, 2016.
\newblock URL \url{http://www.gurobi.com}.

\bibitem[Hong et~al.(2016)Hong, Oh, Lee, and Han]{hong2016learning}
Seunghoon Hong, Junhyuk Oh, Honglak Lee, and Bohyung Han.
\newblock Learning transferrable knowledge for semantic segmentation with deep
  convolutional neural network.
\newblock In \emph{Proceedings of the IEEE Conference on Computer Vision and
  Pattern Recognition}, pages 3204--3212, 2016.

\bibitem[Jean et~al.(2015)Jean, Cho, Memisevic, and Bengio]{jean2014using}
S{\'e}bastien Jean, Kyunghyun Cho, Roland Memisevic, and Yoshua Bengio.
\newblock On using very large target vocabulary for neural machine translation.
\newblock 2015.

\bibitem[Kingma and Ba(2015)]{kingma2014adam}
Diederik~P Kingma and Jimmy Ba.
\newblock Adam: A method for stochastic optimization.
\newblock In \emph{International Conference on Machine Learning}, 2015.

\bibitem[Kirkpatrick et~al.(1983)Kirkpatrick, Gelatt, and
  Vecchi]{kirkpatrick1983optimization}
Scott Kirkpatrick, C~Daniel Gelatt, and Mario~P Vecchi.
\newblock Optimization by simulated annealing.
\newblock \emph{science}, 220\penalty0 (4598):\penalty0 671--680, 1983.

\bibitem[Laporte(1992)]{laporte1992vehicle}
Gilbert Laporte.
\newblock The vehicle routing problem: An overview of exact and approximate
  algorithms.
\newblock \emph{European journal of operational research}, 59\penalty0
  (3):\penalty0 345--358, 1992.

\bibitem[Laporte et~al.(2000)Laporte, Gendreau, Potvin, and
  Semet]{laporte2000classical}
Gilbert Laporte, Michel Gendreau, Jean-Yves Potvin, and Fr{\'e}d{\'e}ric Semet.
\newblock Classical and modern heuristics for the vehicle routing problem.
\newblock \emph{International transactions in operational research}, 7\penalty0
  (4-5):\penalty0 285--300, 2000.

\bibitem[Luong et~al.(2015)Luong, Pham, and Manning]{luong2015effective}
Minh-Thang Luong, Hieu Pham, and Christopher~D Manning.
\newblock Effective approaches to attention-based neural machine translation.
\newblock \emph{Conference on Empirical Methods in Natural Language
  Processing}, 2015.

\bibitem[Mnih et~al.(2015)Mnih, Kavukcuoglu, Silver, Rusu, Veness, Bellemare,
  Graves, Riedmiller, Fidjeland, Ostrovski, et~al.]{mnih2015human}
Volodymyr Mnih, Koray Kavukcuoglu, David Silver, Andrei~A Rusu, Joel Veness,
  Marc~G Bellemare, Alex Graves, Martin Riedmiller, Andreas~K Fidjeland, Georg
  Ostrovski, et~al.
\newblock Human-level control through deep reinforcement learning.
\newblock \emph{Nature}, 518\penalty0 (7540):\penalty0 529--533, 2015.

\bibitem[Mnih et~al.(2016)Mnih, Badia, Mirza, Graves, Lillicrap, Harley,
  Silver, and Kavukcuoglu]{mnih2016asynchronous}
Volodymyr Mnih, Adria~Puigdomenech Badia, Mehdi Mirza, Alex Graves, Timothy
  Lillicrap, Tim Harley, David Silver, and Koray Kavukcuoglu.
\newblock Asynchronous methods for deep reinforcement learning.
\newblock In \emph{International Conference on Machine Learning}, pages
  1928--1937, 2016.

\bibitem[Neubig(2017)]{neubig2017neural}
Graham Neubig.
\newblock Neural machine translation and sequence-to-sequence models: A
  tutorial.
\newblock \emph{arXiv preprint arXiv:1703.01619}, 2017.

\bibitem[Ritzinger et~al.(2016)Ritzinger, Puchinger, and
  Hartl]{ritzinger2016survey}
Ulrike Ritzinger, Jakob Puchinger, and Richard~F Hartl.
\newblock A survey on dynamic and stochastic vehicle routing problems.
\newblock \emph{International Journal of Production Research}, 54\penalty0
  (1):\penalty0 215--231, 2016.

\bibitem[Snyder and Shen(2018)]{snyder2018fundamentals}
Lawrence~V Snyder and Zuo-Jun~Max Shen.
\newblock \emph{Fundamentals of Supply Chain Theory}.
\newblock John Wiley \& Sons, 2nd edition, 2018.

\bibitem[Sutskever et~al.(2014)Sutskever, Vinyals, and
  Le]{sutskever2014sequence}
Ilya Sutskever, Oriol Vinyals, and Quoc~V Le.
\newblock Sequence to sequence learning with neural networks.
\newblock In \emph{Advances in neural information processing systems}, pages
  3104--3112, 2014.

\bibitem[Toth and Vigo(2002)]{toth2002vehicle}
Paolo Toth and Daniele Vigo.
\newblock \emph{The Vehicle Routing Problem}.
\newblock SIAM, 2002.

\bibitem[Vinyals et~al.(2015)Vinyals, Fortunato, and
  Jaitly]{vinyals2015pointer}
Oriol Vinyals, Meire Fortunato, and Navdeep Jaitly.
\newblock Pointer networks.
\newblock In \emph{Advances in Neural Information Processing Systems}, pages
  2692--2700, 2015.

\bibitem[Vinyals et~al.(2016)Vinyals, Bengio, and Kudlur]{vinyals2015order}
Oriol Vinyals, Samy Bengio, and Manjunath Kudlur.
\newblock Order matters: Sequence to sequence for sets.
\newblock 2016.

\bibitem[Voudouris and Tsang(1999)]{voudouris1999guided}
Christos Voudouris and Edward Tsang.
\newblock Guided local search and its application to the traveling salesman
  problem.
\newblock \emph{European journal of operational research}, 113\penalty0
  (2):\penalty0 469--499, 1999.

\bibitem[Williams and Peng(1991)]{williams1991function}
Ronald~J Williams and Jing Peng.
\newblock Function optimization using connectionist reinforcement learning
  algorithms.
\newblock \emph{Connection Science}, 3\penalty0 (3):\penalty0 241--268, 1991.

\bibitem[Wren and Holliday(1972)]{wren1972computer}
Anthony Wren and Alan Holliday.
\newblock Computer scheduling of vehicles from one or more depots to a number
  of delivery points.
\newblock \emph{Operational Research Quarterly}, pages 333--344, 1972.

\bibitem[Xiao et~al.(2015)Xiao, Xu, Yang, Zhang, Peng, and
  Zhang]{xiao2015application}
Tianjun Xiao, Yichong Xu, Kuiyuan Yang, Jiaxing Zhang, Yuxin Peng, and Zheng
  Zhang.
\newblock The application of two-level attention models in deep convolutional
  neural network for fine-grained image classification.
\newblock In \emph{Proceedings of the IEEE Conference on Computer Vision and
  Pattern Recognition}, pages 842--850, 2015.

\bibitem[Xu et~al.(2015)Xu, Ba, Kiros, Cho, Courville, Salakhudinov, Zemel, and
  Bengio]{xu2015show}
Kelvin Xu, Jimmy Ba, Ryan Kiros, Kyunghyun Cho, Aaron Courville, Ruslan
  Salakhudinov, Rich Zemel, and Yoshua Bengio.
\newblock Show, attend and tell: Neural image caption generation with visual
  attention.
\newblock In \emph{International Conference on Machine Learning}, pages
  2048--2057, 2015.

\end{thebibliography}

\newpage\clearpage
\appendix

\section{Our Model versus Pointer Network}\label{sec:comp-with-pointer}

In this section we use  the \textit{Traveling Salesman Problem} (TSP) (a special case of the VRP in which there is only a single route to optimize) as the test-bed to validate the performance of the proposed method. We compare the route lengths of the TSP solutions obtained by our framework with those given by the model of \citet{bello2016neural} for random instances with 20, 50, and 100 nodes. In the training phase, we generate $10^6$ TSP instances for each problem size, and use them in training for 20 epochs. $10^6$ is chosen because we want to have a diverse set of problem configurations; it can be larger or smaller, or we can generate the instances on-the-fly as long as we make sure that the instance are drawn from the same probability distribution with the same random seed. The city locations are chosen uniformly from the unit square $[0,1]\times [0,1]$. We use the same data distribution to produce instances for the testing phase. The decoding process starts from a random TSP node and the termination criterion is that all cities are visited. We also use a masking scheme to prohibit visiting nodes more than once.

Table \ref{table:tsp-path} summarizes the results for different TSP sizes using the \textit{greedy decoder} in which at every decoding step, the city with the highest probability is chosen as the destination. The results are averaged over 1000 instances. The first column is the average TSP tour length using our proposed architecture, the second column is the result of our implementation of \citet{bello2016neural} with greedy decoder, and the optimal tour lengths are reported in the last column. To obtain the optimal values, we solved the TSP using the Concorde optimization software \cite{applegate2006traveling}. A comparison of the first two columns suggests that there is almost no difference between the performance of our framework and Pointer-RL. In fact, the RNN encoder of the Pointer Network learns to convey no information to the next steps, i.e., $h_t = f(x_t)$. On the other hand, our approach is around 60\% faster in both training and inference, since it has two fewer RNNs---one in the encoder of actor network and another in the encoder of critic network. Table \ref{table:tsp-path} also summarizes the training times for one epoch of the training and the time-savings that we gain by eliminating the encoder RNNs.

\begin{table}[htbp]
	\centering
	\caption{Average tour length for TSP and training time for one epoch (in minutes).}
	\label{table:tsp-path}
	\scalebox{0.8 }{
		\begin{tabular}{l m{2.2cm} m{1.7cm} m{1.7cm} m{2.2cm} m{1.7cm} m{2.2cm} }
			\toprule
			& \multicolumn{3}{c}{Average tour length }&\multicolumn{3}{c}{Training time }\\
			\cmidrule(lr){2-4} \cmidrule(lr){5-7}
			Task  & Our Framework (Greedy) & Pointer-RL (Greedy) & Optimal & Our Framework (Greedy) & Pointer-RL (Greedy) & \% Time Saving \\\midrule
			TSP20      &3.97& 3.96 &3.84&22.18& 50.33 & 55.9\%\\
			TSP50   & 6.08 &6.05 &5.70& 54.10 &147.25 & 63.3\%\\
			TSP100   & 8.44 &8.45 &7.77& 122.10 &300.73 & 59.4\%\\
			\bottomrule
	\end{tabular}}
\end{table}

\section{Capacitated VRP Baselines}\label{sec:vrp:heuristic}
In this Appendix, we briefly describe the algorithms and solvers that we used as benchmarks. More details and examples of these algorithms can be found in \citet{snyder2018fundamentals}. The first two baseline approaches are well-known heuristics designed for VRP. Our third baseline is Google's optimization tools, which includes one of the best open-source VRP solvers. Finally, we compute the optimal solutions for small VRP instances, so we can measure how far the solutions are from optimality.

\subsection{Clarke-Wright Savings Heuristic} \label{sec:vrp:heuristic:clarke}
The Clarke-Wright savings heuristic \cite{clarke1964scheduling} is one of the best-known heuristics for the VRP. Let $\mathcal{N} \doteq \{1,\cdots,N\}$ be the set of customer nodes, and 0 be the depot. The distance between nodes $i$ and $j$ is denoted by $c_{ij}$, and $c_{0i}$ is the distance of customer $i$ from the depot. Algorithm \ref{alg:vrp:clarke} describes a randomized version of the heuristic. The basic idea behind this algorithm is that it initially considers a separate route for each customer node $i$, and then reduces the total cost by iteratively merging the routes. Merging two routes by adding the edge $(i,j)$ reduces the total distance by $s_{ij}=c_{i0} + c_{0j} - c_{ij}$, so the algorithm prefers mergers with the highest savings $s_{ij}$. 

We introduce two hyper-parameters, $R$ and $M$, which we refer to as the \textit{randomization depth} and \textit{randomization iteration}, respectively.
When $M=R=1$, this algorithm is equivalent to the original Clarke-Wright savings heuristic, in which case, the feasible merger with the highest savings will be selected. By allowing $M,R>1$, we introduce randomization, which can improve the performance of the algorithm further. In particular, Algorithm \ref{alg:vrp:clarke} chooses randomly from the $r\in\{1,\cdots,R\}$ best feasible mergers. Then, for each $r$, it solves the problem $m\in\{1,\cdots,M\}$ times, and returns the solution with the shortest total distance. 

\begin{algorithm}
	\caption{Randomized Clarke-Wright Savings Heuristic}
	\label{alg:vrp:clarke}
	\begin{algorithmic}[1]
		\STATE compute savings $s_{ij}$, where  
		\vspace{-.3cm}
		\begin{align}
		s_{ij} &= c_{i0} + c_{0j} - c_{ij} &\quad i,j\in\mathcal{N}, \,i\neq j \nonumber\\
		s_{ii} &= 0  &i\in\mathcal{N}\nonumber
		\end{align}
		\vspace{-.7cm}
		\FOR{$r = 1,\cdots,R$}
		\FOR{$m = 1,\cdots,M$}
		\STATE place each $i\in\mathcal{N}$ in its own route
		\REPEAT
		\STATE find $k$ feasible mergers $(i,j)$ with the highest $s_{ij}>0$, satisfying the following conditions:
		
		{\setlength{\leftskip}{.5cm}
			i) $i$ and $j$ are  in different routes\\
			ii) both $i$ and $j$ are adjacent to the depot\\
			iii) combined demand of routes containing $i$ and $j$ is $\leq$ vehicle capacity\\
		}
		
		\STATE choose a random $(i,j)$ from the feasible mergers, and combine the associated routes by replacing $(i,0)$ and $(0,j)$ with $(i,j)$
		\UNTIL{no feasible merger is left}
		
		\ENDFOR
		\ENDFOR
		\STATE\textbf{Return}: route with the shortest length
	\end{algorithmic}
\end{algorithm}

\subsection{Sweep Heuristic}
The sweep heuristic \cite{wren1972computer} solves the VRP by breaking it into multiple TSPs. By rotating an arc emanating from the depot, it groups the nodes into several clusters, while ensuring that the total demand of each cluster is not violating the vehicle capacity. Each cluster corresponds to a TSP that can be solved by using an exact or approximate algorithm. In our experiments, we use dynamic programming to find the optimal TSP tour. After solving TSPs, the VRP solution can be obtained by combining the TSP tours. Algorithm \ref{alg:vrp:sweep} shows the pseudo-code of this algorithm.

\begin{algorithm}[h]
	\caption{Randomized Sweep Algorithm}
	\label{alg:vrp:sweep}
	\begin{algorithmic}[1]
		\STATE for each $i\in\mathcal{N}$, compute angle $\alpha_i$, respective to depot location
		\STATE $l\leftarrow$ vehicle capacity 
		\FOR{$r=1,\cdots,R$}
		\STATE select a random angle $\alpha$
		\STATE $k\leftarrow 0$; initialize cluster $S_k\leftarrow\emptyset$
		\REPEAT
		\STATE increase $\alpha$ until it equal to some $\alpha_i$
		\IF{demand $d_i > l$}
		\STATE $k \leftarrow k+1$
		\STATE $S_k\leftarrow\emptyset$
		\STATE $l\leftarrow$ vehicle capacity 
		\ENDIF
		\STATE $S_k \leftarrow S_k \cup \{i\}$
		\STATE $l\leftarrow l-d_i$ 
		
		\UNTIL{no unclustered node is left}
		\STATE solve a TSP for each $S_k$
		\STATE merge TSP tours to produce a VRP route
		\ENDFOR
		\STATE \textbf{Return}: route with the shortest length
		
	\end{algorithmic}
\end{algorithm}

\subsection{Google's OR-Tools}
Google Optimization Tools (OR-Tools) \cite{ortools} is an open-source solver for combinatorial optimization problems. OR-Tools contains one of the best available VRP solvers, which has implemented many heuristics (e.g., Clarke-Wright savings heuristic \cite{clarke1964scheduling}, Sweep heuristic \cite{wren1972computer}, Christofides' heuristic \cite{christofides1976worst} and a few others) for finding an initial solution and metaheuristics (e.g. Guided Local Search \cite{voudouris1999guided}, Tabu Search \cite{glover2013tabu} and Simulated Annealing \cite{kirkpatrick1983optimization}) for escaping from  local minima in the search for the best solution. The default version of the OR-Tools VRP solver does not exactly match the VRP studied in this paper, but with a few adjustments, we can use it as our baseline. The first limitation is that OR-Tools only accepts integer locations for the customers and depot while our problem is defined on the unit square. To handle this issue, we scale up the problem by multiplying all locations by $10^4$ (meaning that we will have 4 decimal digits of precision), so the redefined problem is now in $[0,10^4]\times[0,10^4]$. After solving the problem, we scale down the solutions and tours to get the results for the original problem. The second difference is that OR-Tools is defined for a VRP with multiple vehicles, each of which can have at most one tour. One can verify that by setting a large number of vehicles (10 in our experiments), it is mathematically equivalent to our version of the VRP.

\subsection{Optimal Solution}
We use a mixed integer formulation for the VRP \cite{toth2002vehicle} and the Gurobi optimization solver \cite{gurobi} to obtain the optimal VRP tours. VRP has an exponential number of constraints, and of course, it requires careful tricks for even small problems. In our implementation, we start off with a relaxation of the capacity constraints and solve the resulting problem to obtain a lower bound on the optimal objective value. Then we check the generated tours and add the capacity constraint as \textit{lazy-constraints} if a specific subtour's demand has violated the vehicle capacity, or the subtour does not include the depot. With this approach, we were able to find the optimal solutions for VRP10 and VRP20, but this method is intractable for larger VRPs; for example, on a single instance of VRP50, the solution has 6.7\% optimality gap after 10000 seconds.

\section{Extended Results of the VRP Experiment}
In this section, we present more detailed results for the VRP, including a comparison with baselines and an illustration of the  solutions generated. We demonstrate the flexibility of the model to incorporate split deliveries, as an option, to further improve the solution quality. We also illustrate with an example that our proposed framework can be applied to more challenging VRPs with the stochastic elements. 

\subsection{Implementation Details}
For the embedding, we use 1-dimensional convolution layers for the embedding, in which the in-width is the input length, the number of filters is $D$, and the number of in-channels is the number of elements of $x$. We find that training without an embedding layer always yields an inferior solution. One possible explanation is that the policy is able to extract useful features from the high-dimensional input representations much more efficiently. Recall that our embedding is an affine transformation, so it does not necessarily keep the embedded input distances proportional to the original 2-dimensional Euclidean distances.

We use one layer of LSTM RNN in the decoder with a state size of 128. Each customer location is also embedded into a vector of size 128, shared among the inputs. We employ similar embeddings for the dynamic elements; the demand $d^i_t$ and the remaining vehicle load after visiting node $i$, $l_t-d^i_t$, are mapped to a vector in a 128-dimensional vector space and used in the attention layer. In the critic network, first, we use the output probabilities of the actor network to compute a weighted sum of the embedded inputs, and then, it has two hidden layers: one dense layer with ReLU activation and another linear one with a single output. The variables in both actor and critic network are initialized with Xavier initialization \cite{glorot2010understanding}. For training both networks, we use the REINFORCE Algorithm and Adam optimizer \cite{kingma2014adam} with learning rate $10^{-4}$. The batch size $N$ is 128, and we clip the gradients when their norm is greater than 2. We use dropout with probability 0.1 in the decoder LSTM. Moreover, we tried the entropy regularizer \cite{williams1991function,mnih2016asynchronous}, which has been shown to be useful in preventing the algorithm from getting stuck in local optima, but it does not show any improvement in our experiments; therefore, we do not use it in the results reported in this paper.

On a single GPU K80, every 100 training steps of the VRP with 20 customer nodes takes approximately 35 seconds. Training for 20 epochs requires about 13.5 hours. The TensorFlow implementation of our code will be publicly available.

\subsection{Flexibility to VRPs with Split Demands}\label{sec:split-demand}
In the classical VRP that we studied in Section \ref{sec:vrp-experiment}, each customer
is required to be visited exactly once. On the contrary to what is usually assumed in the classical VRP, one can relax this constraint to obtain savings by allowing split deliveries \cite{archetti2008split}. %Assuming the integrality of the demands that we consider in our experiments, this problem is much more complicated than ``traditional'' VRP since one needs to introduce many new integer variables to the mixed integer formulation. 
In this section, we show that this relaxation is straightforward by slightly modifying the masking scheme. Basically, we omit the condition \textit{(iii)} from the masking introduced in Section \ref{sec:vrp-experiment}, and use the new masking with the exactly similar model; we want to emphasize that we do not re-train the model and use the variables trained previously, so this property is achieved at no extra cost. 

\begin{figure*}[h]
	\centering
	\begin{subfigure}{.45\columnwidth}
		\includegraphics[width=\columnwidth]{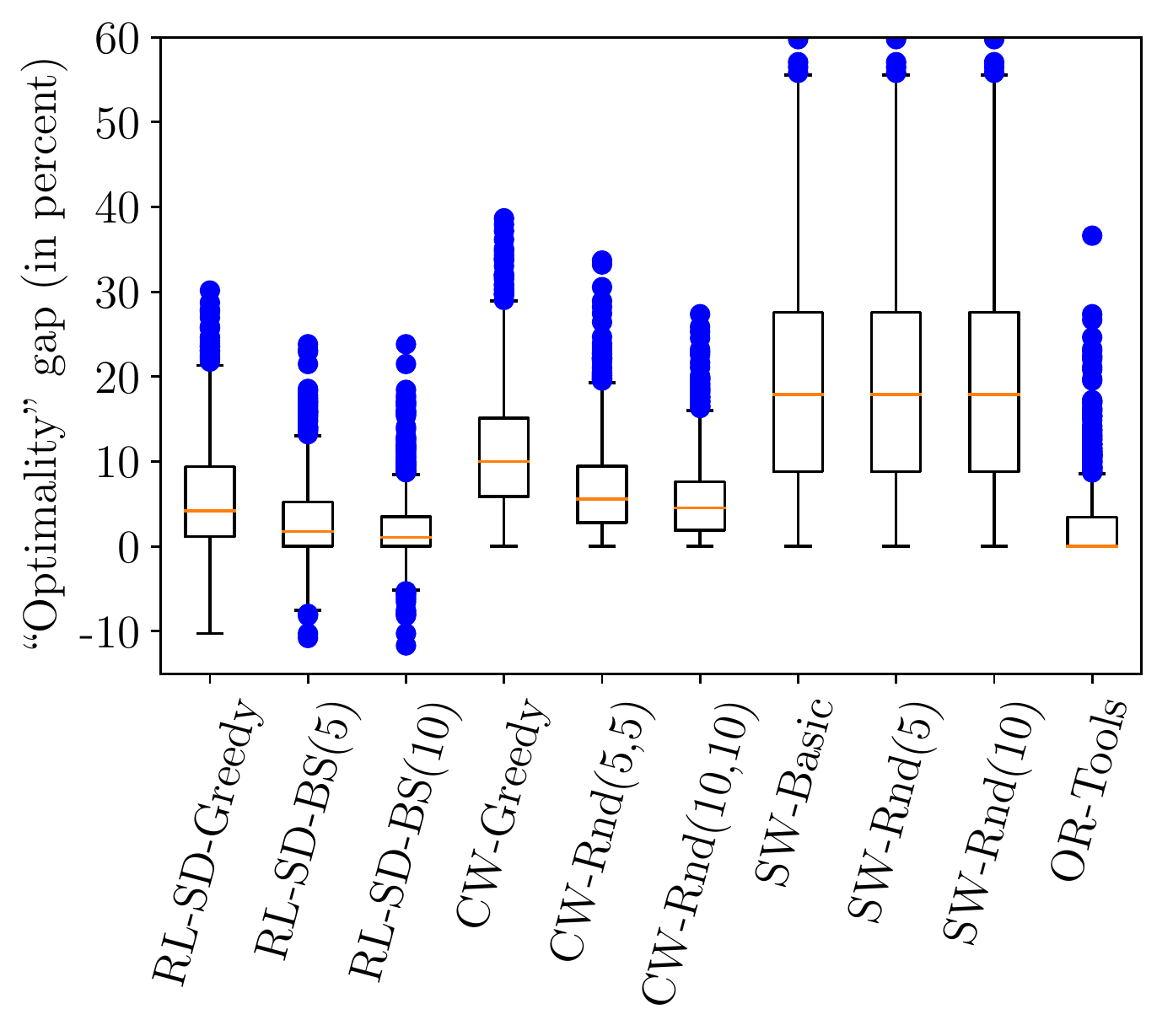}
		\caption{Comparison for VRP10}
		\label{vrp:sp:gap:compa}
	\end{subfigure}
	\begin{subfigure}{.45\columnwidth}
		\includegraphics[width=\columnwidth]{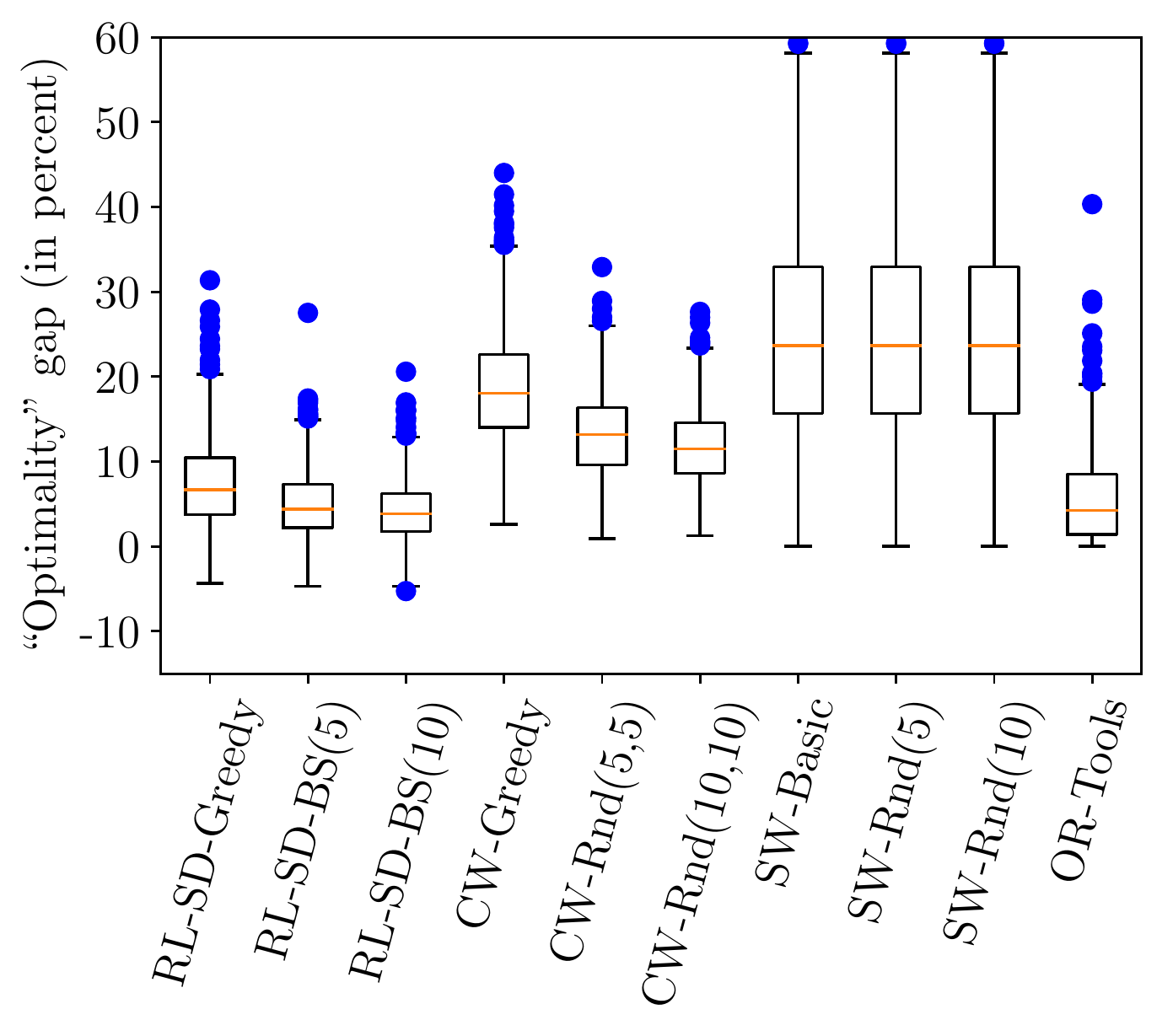}
		\caption{Comparison for VRP20}
		\label{vrp:sp:gap:compb}
	\end{subfigure}
	\begin{subfigure}{.49\columnwidth}
		\includegraphics[width=1.05\columnwidth]{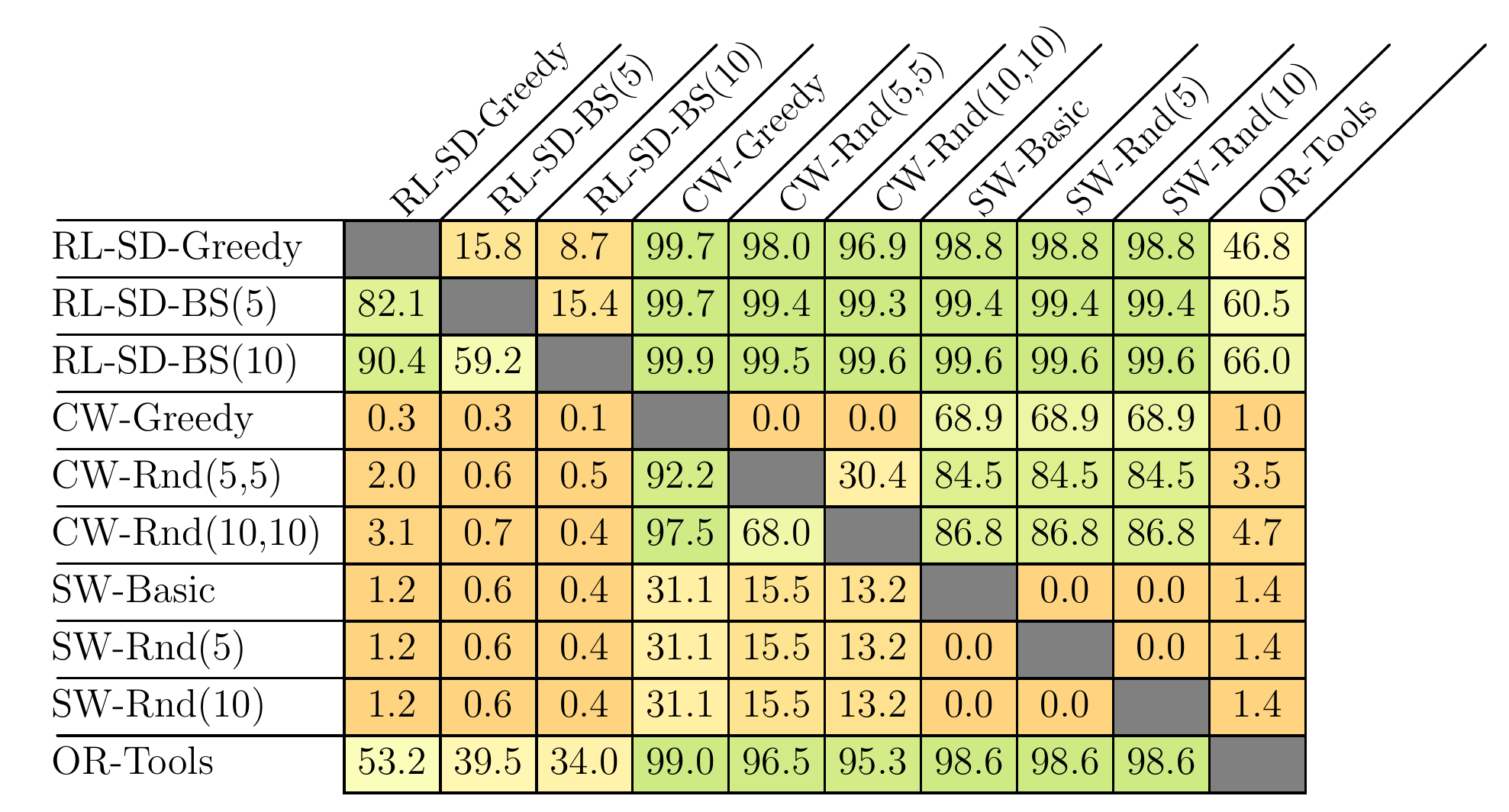}
		\caption{Comparison for VRP50}
		\label{vrp:sp:frac:compc}
	\end{subfigure}
	\begin{subfigure}{.49\columnwidth}
		\includegraphics[width=1.05\columnwidth]{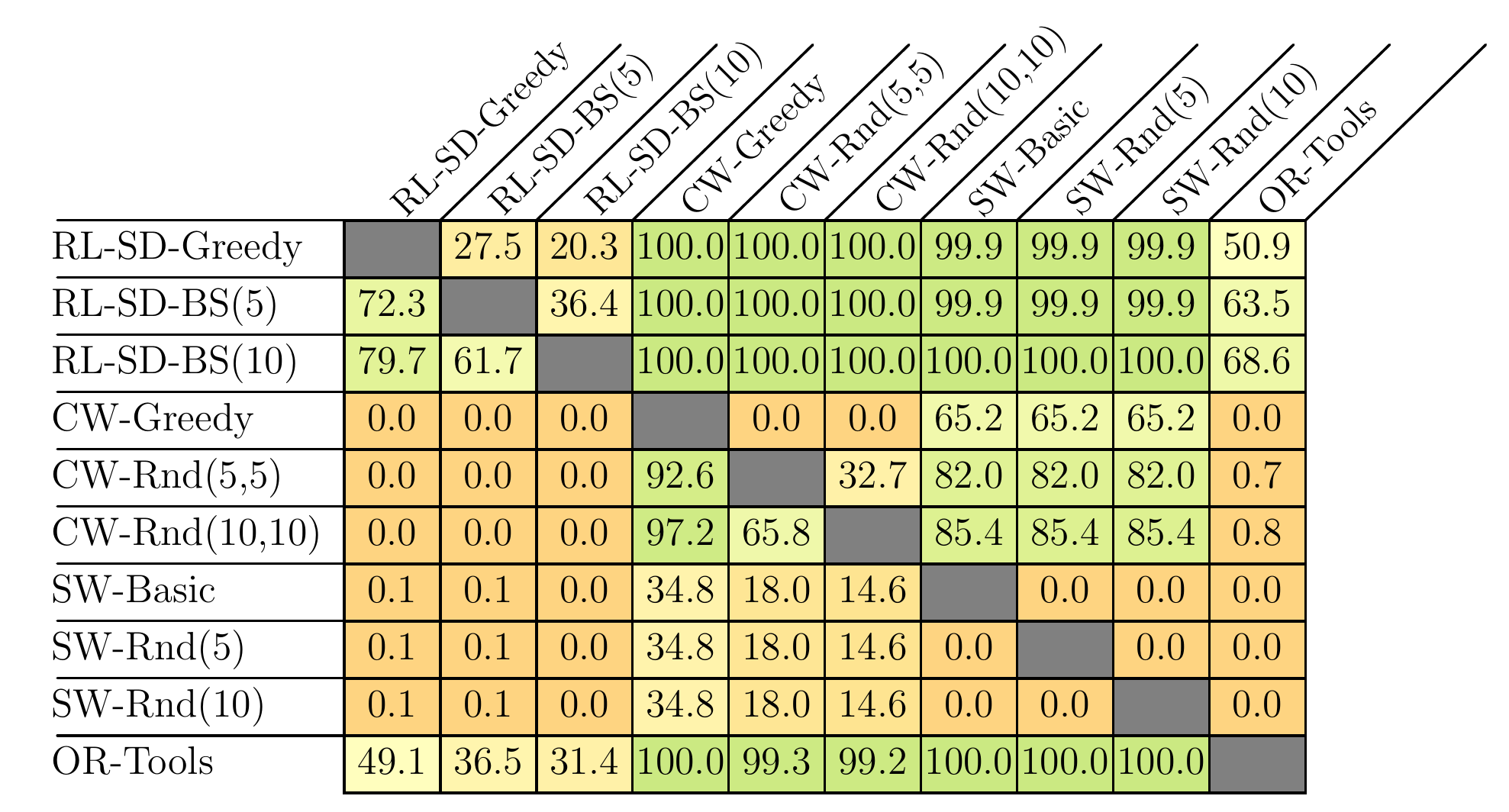}
		\caption{Comparison for VRP100}
		\label{vrp:sp:frac:compd}
	\end{subfigure}
	
	\caption{Parts \ref{vrp:gap:compa} and \ref{vrp:gap:compb} show the ``optimality'' gap (in percent) using different algorithms/solvers for VRP10 and VRP20. Parts \ref{vrp:frac:compc} and \ref{vrp:frac:compd} give the proportion of the samples (in percent) for which the algorithms in the rows outperform those in the columns; for example, RL-BS(5) is provides shorter tours compared to RL-greedy in 82.1\% of the VRP50 instances. }
	\label{fig:sp:vrp:cvrp-comp}
\end{figure*}

Figure \ref{fig:sp:vrp:cvrp-comp} shows the improvement by relaxing these constraints, where we label our relaxed method with ``RL-SD''. Other heuristics does not have such option and they are reported for the original (not relaxed) problem. In parts \ref{vrp:sp:gap:compa} and \ref{vrp:sp:gap:compb} we illustrate the ``optimality'' gap for VRP10 and VRP20, respectively. What we refer to optimality in this section (and other places in this paper) is the optimal objective value of the non-relaxed problem. Of course, the relaxed problem would have a lower optimal objective value. That is why RL-SD obtains negative values in these plots. We see that RL-SD can effectively use split delivery to obtain solutions that are around $5-10\%$ shorter than the ``optimal'' tours. Similar to \ref{fig:vrp:cvrp-comp}, parts \ref{vrp:sp:frac:compc} and \ref{vrp:sp:frac:compd} show the winning percentage of the algorithms in rows in comparison to the ones in columns. We observe that the winning percentage of RL-SD methods significantly improves after allowing the split demands. For example in VRP50 and VRP100, RL-SD-Greedy is providing competitive results with OR-Tools, or RL-SD-BS(10) outperforms OR-Tools in roughly $67\%$ of the instances, while this number was around $61\%$ before relaxation.

\subsection{Summary of Comparison with Baselines}
Table \ref{table:cvrp-path-time} provides the average and the standard deviation of tour lengths for different VRPs. We also test the RL approach using the split delivery option where the customer demands can be satisfied in more than one subtours (labeled with ``RL-SD'', at the end of the table).  We observe that the average total length of the solutions found by our method using various decoders outperforms the heuristic algorithms and OR-Tools. We also see that using the beam search decoder significantly improves the solution while only adding a small computational cost in run-time. Also allowing split delivery enables our RL-based methods to improve the total tour length by a factor of around $0.6\%$ on average. We also present the solution time comparisons in this table, where all the times are reported on a single core Intel 2.6 GHz CPU.

\begin{table}[htbp]
	\centering
	\caption[Average total tour length]{Average tour length, standard deviations of the tours and the average solution time (in seconds) using different baselines over a test set of size 1000.}
	\label{table:cvrp-path-time}
	\resizebox{\textwidth}{!}{
		\begin{tabular}{lcccccccccccc}
			\toprule
			\multirow{2}{*}{Baseline}&\multicolumn{3}{c}{VRP10, Cap20} &\multicolumn{3}{c}{VRP20, Cap30} &\multicolumn{3}{c}{VRP50, Cap40} &\multicolumn{3}{c}{VRP100, Cap50}
			\\\cmidrule(lr){2-4}\cmidrule(lr){5-7}\cmidrule(lr){8-10}\cmidrule(lr){11-13}
			
			& mean& std & time & mean& std & time & mean& std & time & mean& std & time\\\midrule
			
			RL-Greedy &4.84 &0.85 &0.049 &6.59 &0.89 &0.105 &11.39 &1.31 & 0.156 &17.23 &1.97 &0.321 \\
			RL-BS(5)  &4.72 &0.83 &0.061 &6.45 &0.87 &0.135 &11.22 &1.29 &0.208 &17.04 &1.93 & 0.390\\
			RL-BS(10) &4.68 &0.82 &0.072 &\textbf{6.40} &0.86 &0.162 &\textbf{11.15} &1.28 &0.232 &\textbf{16.96} &1.92 &0.445 \\\midrule

			CW-Greedy    &5.06 &0.85 &0.002 &7.22 &0.90  &0.011 &12.85 &1.33 &0.052 &19.72 &1.92 &0.186 \\
			CW-Rnd(5,5)  &4.86 &0.82 &0.016 &6.89 &0.84  &0.053 &12.35 &1.27 &0.217 &19.09 &1.85 &0.735 \\
			CW-Rnd(10,10)&4.80 &0.82 &0.079 &6.81 &0.82  &0.256 &12.25 &1.25 &0.903 &18.96 &1.85 &3.171 \\\midrule
			SW-Basic     &5.42 &0.95 &0.001 &7.59 &0.93  &0.006 &13.61 &1.23 &0.096 &21.01 &1.51 &1.341 \\
			SW-Rnd(5)    &5.07 &0.87 &0.004 &7.17 &0.85  &0.029 &13.09 &1.12 &0.472 &20.47 &1.41 &6.32 \\
			SW-Rnd(10)   &5.00 &0.87 &0.008 &7.08 &0.84  &0.062 &12.96 &1.12 &0.988 &20.33 &1.39 &12.443 \\\midrule
			OR-Tools     &\textbf{4.67} &0.81 &0.004 &6.43 &0.86  &0.010 &11.31 &1.29 &0.053 &17.16 &1.88 &0.231 \\\midrule
			Optimal      &4.55 &0.78 &0.029 &6.10 &0.79  &102.8 &\multicolumn{3}{c}{\textemdash}  & \multicolumn{3}{c}{\textemdash}\\[1pt]\hline\midrule
	
			RL-SD-Greedy    &4.80 &0.83 &0.059 &6.51 &0.84  &0.107 &11.32 &1.27 &0.176 &17.12 &1.90 &0.310 \\
			RL-SD-BS(5)     &4.69 &0.80 &0.063 &6.40 &0.85  &0.145 &11.14 &1.25 &0.226 &16.94 &1.88 &0.401 \\
			RL-SD-BS(10)    &4.65 &0.79 &0.074 &6.34 &0.80  & 0.155&11.08 &1.24 &0.250 &16.86 &1.87 &0.477 \\\bottomrule
	\end{tabular}}
\end{table}

\subsection{Sample VRP Solutions}\label{sec:sample-vrp-solution}
\begin{figure*}[htbp]
\centering
\begin{subfigure}{.32\columnwidth}
\includegraphics[width=\columnwidth]{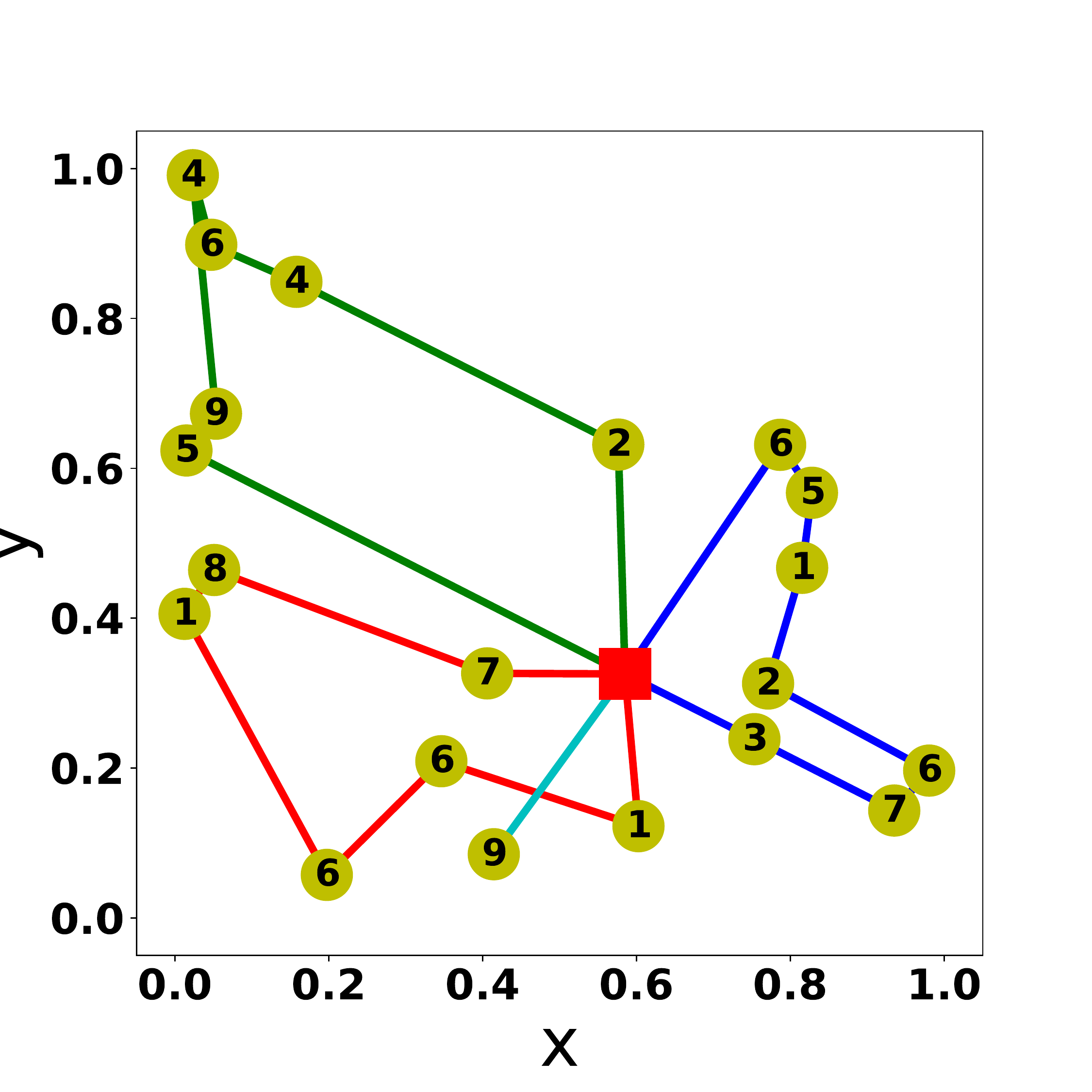}
\caption*{Path length = 5.72}
\includegraphics[width=\columnwidth]{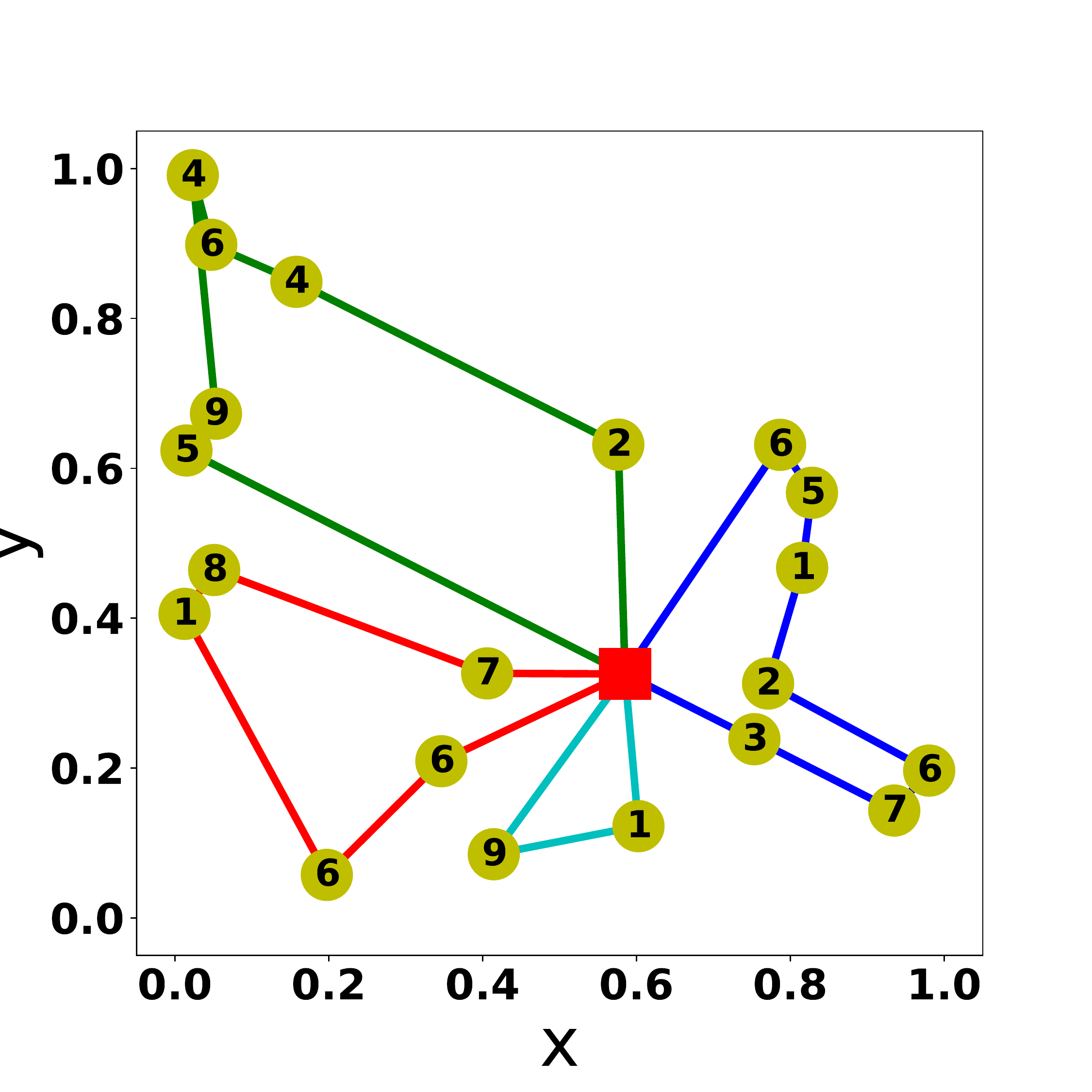}
\caption*{Path length = 5.62}
\caption{Example 1: VRP20;\\ capacity 30}\label{fig:vrp:sample:1}
\end{subfigure}
\begin{subfigure}{.32\columnwidth}
\includegraphics[width=\columnwidth]{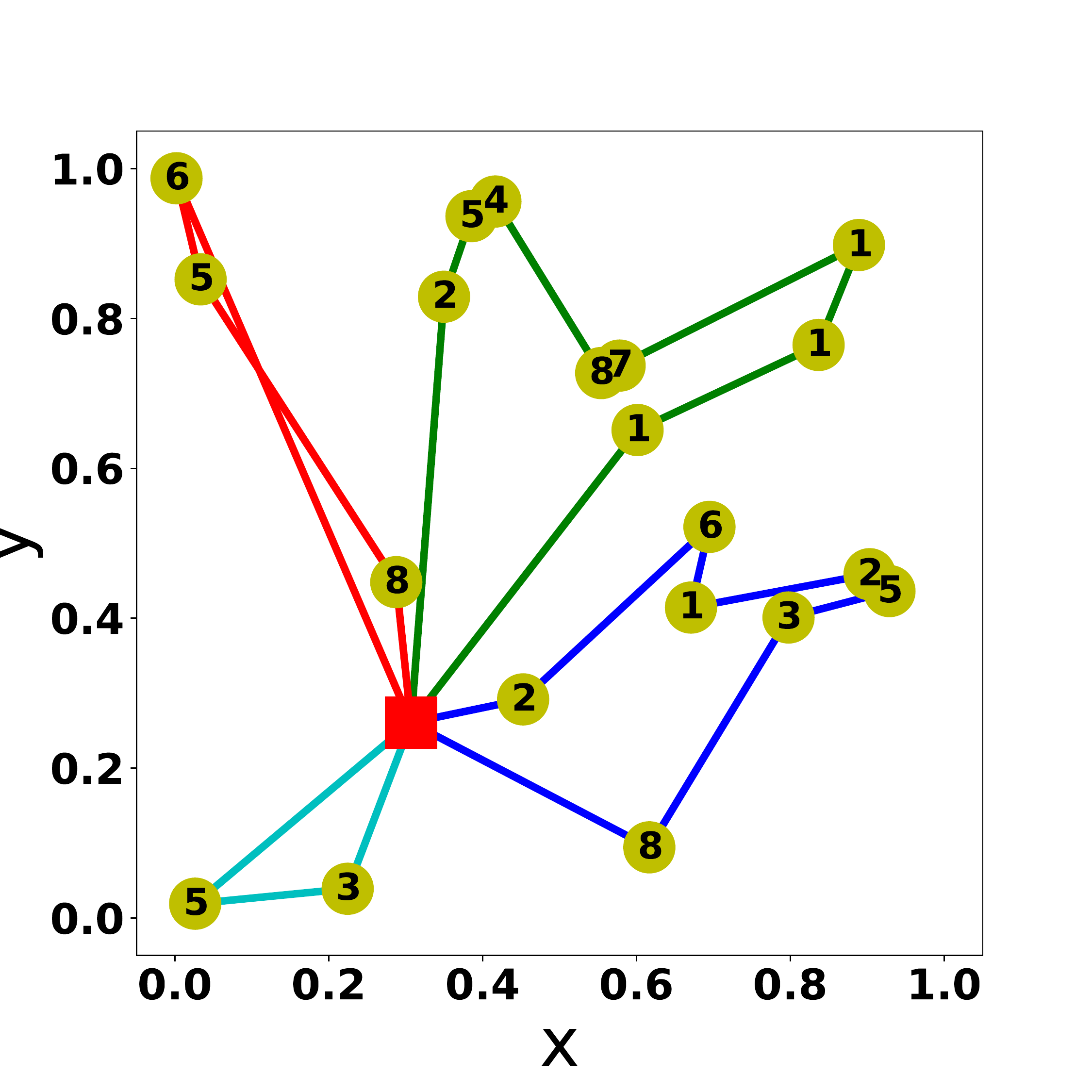}
\caption*{Path length = 6.36}
\includegraphics[width=\columnwidth]{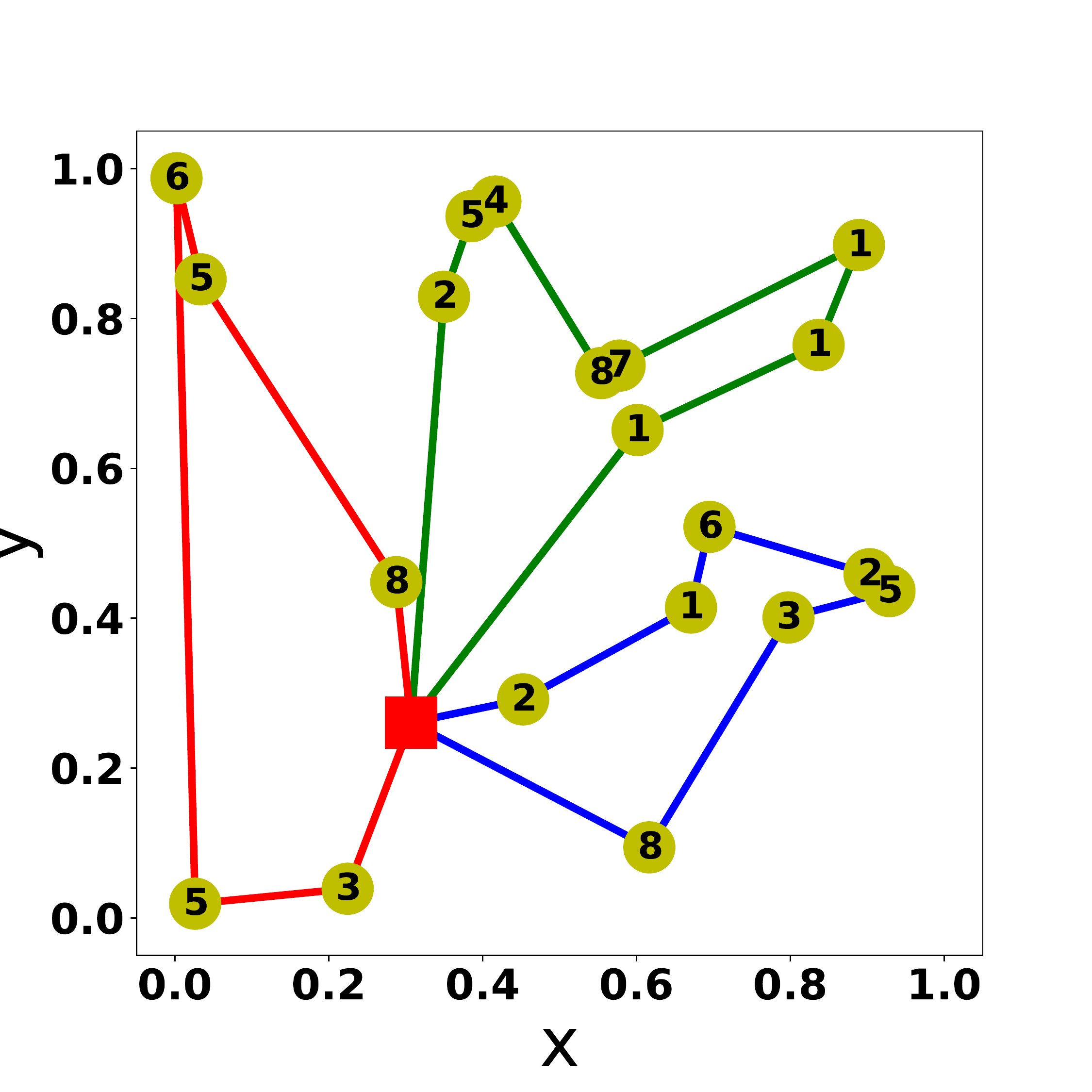}
\caption*{Path length = 6.06}
\caption{Example 2: VRP20;\\ capacity 30}\label{fig:vrp:sample:2}
\end{subfigure}
\begin{subfigure}{.32\columnwidth}
\includegraphics[width=\columnwidth]{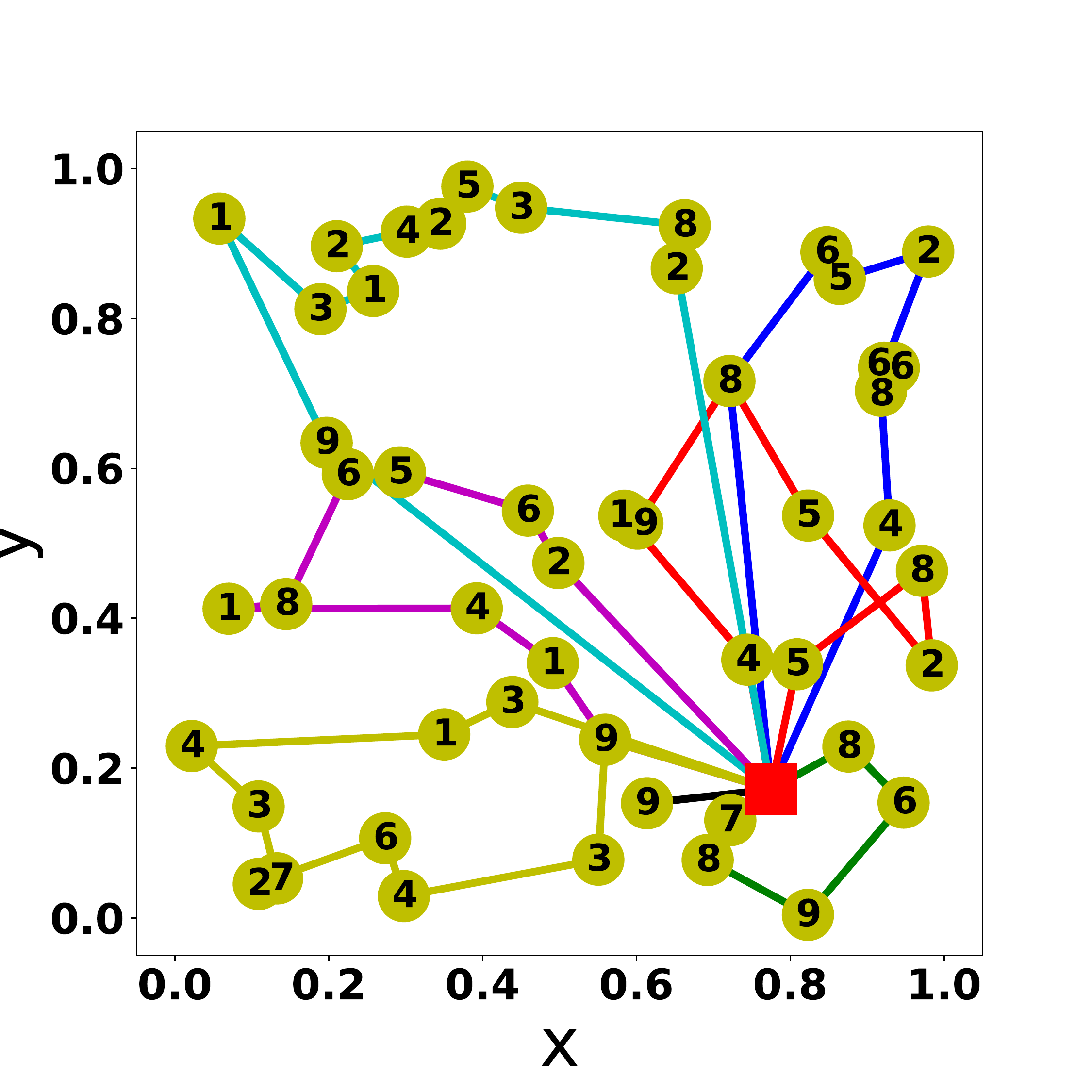}
\caption*{Path length = 10.72}
\includegraphics[width=\columnwidth]{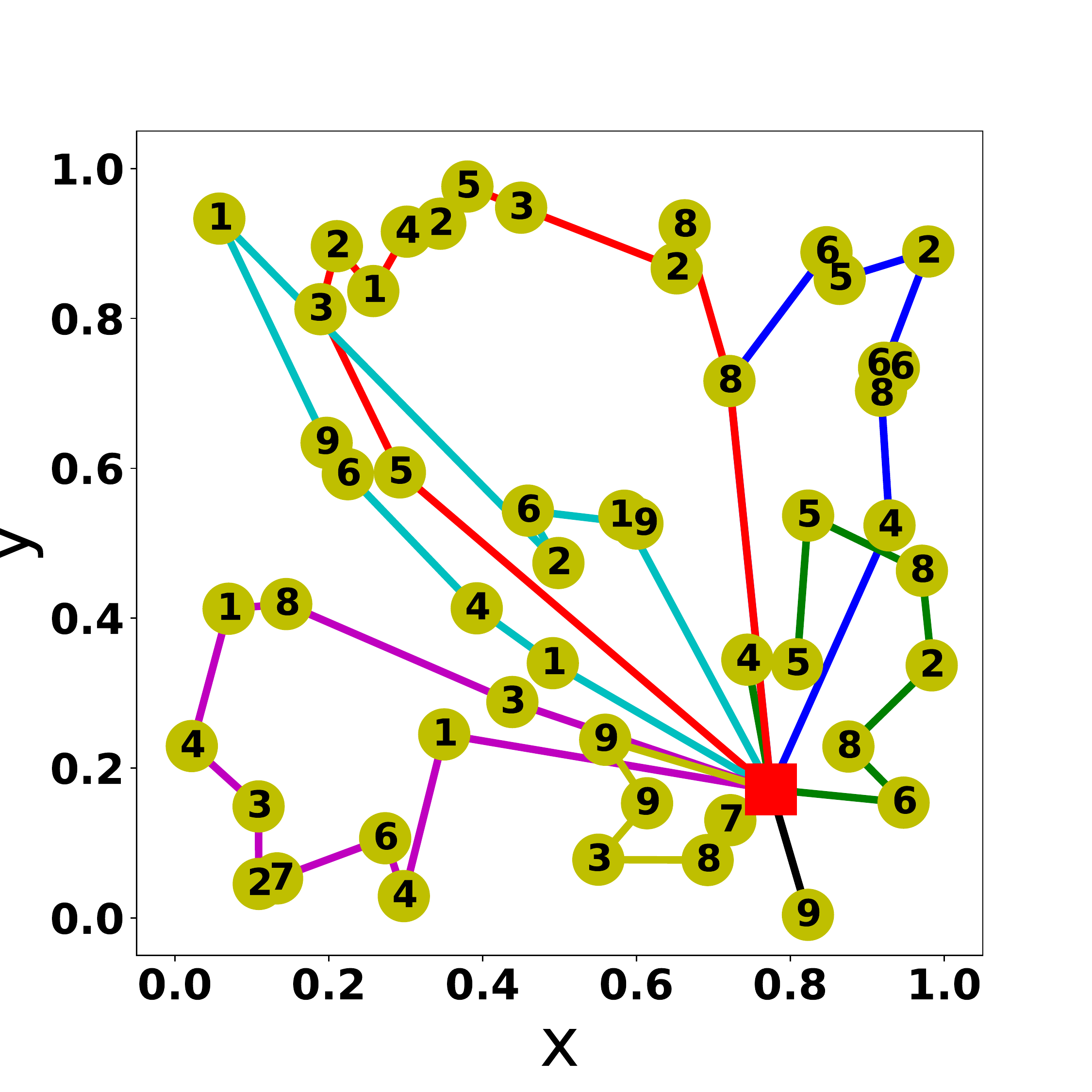}
\caption*{Path length = 10.70}
\caption{Example 3: VRP50;\\ capacity 40}\label{fig:vrp:sample:3}
\end{subfigure}

\caption{Sample decoded solutions for VRP20 and VRP50 using greedy (in top row) and beam-search (bottom row) decoder. The numbers inside the nodes are the demand values. }	
\label{fig:vrp:cvrp-sample}
\end{figure*}

Figure \ref{fig:vrp:cvrp-sample} illustrates sample VRP20 and VRP50 instances decoded by the trained model. The greedy and beam-search decoders were used to produce the figures in the top and bottom rows, respectively.  It is evident that these solutions are not optimal. For example, in part (\subref{fig:vrp:sample:1}), one of the routes crosses itself, which is never optimal in Euclidean VRP instances. Another similar suboptimality is evident in part (\subref{fig:vrp:sample:3}) to make the total distance shorter. However, the figures illustrate how well the policy model has understood the problem structure. It tries to satisfy demands at nearby customer nodes until the vehicle load is small. Then, it automatically comprehends that visiting further nodes is not the best decision, so it returns to the depot and starts a new tour. One interesting behavior that the algorithm has learned can be seen in part (\subref{fig:vrp:sample:3}), in which the solution reduces the cost by making a partial delivery; in this example, we observe that the red and blue tours share a customer node with demand 8, each satisfying a portion of its demand; in this way, we are able to meet all demands without needing to initiate a new tour. We also observe how using the beam-search decoder produces further improvements; for example, as seen in parts (\subref{fig:vrp:sample:2})--(\subref{fig:vrp:sample:3}), it reduces the number of times when a tour crosses itself; or it reduces the number of tours required to satisfy all demands as is illustrated in (\subref{fig:vrp:sample:2}).

Tables \ref{table:vrp:vrp-sample-detail} and \ref{table:vrp:vrp-sample-detail-2} present the solutions found by our model using the greedy and beam search decoders for two sample VRP10 instances with a vehicle capacity of 20. We have 10 customers indexed $0\cdots 9$ and the location with the index 10 corresponds to the depot. The first line specifies the customer locations as well as their demands and the depot location. The solution in the second line is the tour found by the greedy decoder. In the third and fourth line, we observe how increasing the beam width helps in improving the solution quality. Finally, we present the optimal solution in the last row. In \ref{table:vrp:vrp-sample-detail-2}, we illustrate an example where our method has discovered a solution by splitting the demands which is, in fact, considerably shorter than the optimal solution found by solving the mixed integer programming model.

\afterpage{
\begin{table}[ht]
	\centering
	\caption{Solutions found for a sample VRP10 instance. We use different decoders for producing these solutions; the optimal route is also presented in the last row.}
	\label{table:vrp:vrp-sample-detail}
	\scalebox{.9}{
	\begin{tabular}{m{14cm}}
		\toprule
	\textbf{Sample instance for VRP10}:\\
    Customer locations: [[0.411, 0.559], [0.874, 0.302], [0.029, 0.127], [0.188, 0.979], [0.812, 0.330], [0.999, 0.505], [0.926, 0.705], [0.508, 0.739], [0.424, 0.201], [0.314, 0.140]]\\
    Customer demands: [2, 4, 5, 9, 5, 3, 8, 2, 3, 2]\\
    Depot location: [0.890, 0.252]\\

	\midrule

\textbf{Greedy decoder}:\\
Tour Length: 5.305 \\
Tour: 10 $\to$ 5 $\to$ 6 $\to$ 4 $ \to$ 1 $ \to$ 10 $ \to$ 7 $ \to$ 3 $ \to$ 0 $ \to$ 8 $\to$ 9 $\to$ 10 $\to$ 2 $ \to$ 10\\
\midrule
\textbf{BS decoder with width 5}:\\
Beam tour lengths: [5.305, 5.379, 4.807,  5.018, 4.880]\\
Best beam: 2, Best tour length: 4.807\\
Best tour: 10 $\to$ 5 $\to$ 6$\to$ 4 $\to$ 1$\to$ 10 $\to$ 7 $\to$ 3 $\to$ 0$\to$ 10 $\to$ 8 $\to$ 2 $\to$ 9 $\to$ 10\\
\midrule

\textbf{BS decoder with width 10}:\\
Beam tour lengths: [5.305, 5.379, 4.807, 5.0184, 4.880, 4.800, 5.091,
4.757, 4.8034, 4.764]\\
Best beam: 7, Best tour length: 4.757\\
Best tours: 10 $\to$ 5 $\to$ 6 $\to$ 1 $\to$ 10 $\to$ 7 $ \to$ 3 $ \to$ 0 $ \to$ 4 $ \to$ 10 $ \to$ 8 $ \to$ 2 $\to$ 9 $ \to$ 10\\
\midrule
\textbf{Optimal}:\\
Optimal tour length: 4.546\\
Optimal tour: 10 $\to$ 1 $\to$ 10 $\to$ 2 $\to$ 3 $\to$ 8 $\to$ 9 $\to$ 10 $\to$ 0 $\to$ 4 $\to$ 5 $\to$ 6 $\to$ 7 $\to$ 10\\
\bottomrule
	\end{tabular}}
\end{table}
\begin{table}[ht]
	\centering
	\caption{Solutions found for a sample VRP10 instance where by splitting the demands, our method significantly improves upon the ``optimal'' (of which no split demand is allowed).}
	\label{table:vrp:vrp-sample-detail-2}
	\scalebox{.9}{
	\begin{tabular}{m{14cm}}
		\toprule
		\textbf{Sample instance for VRP10}:\\
		Customer locations:  [[0.253, 0.720], [0.289, 0.725], [0.132, 0.131], [0.050, 0.609], [0.780, 0.549], [0.014, 0.920], [0.624, 0.655], [0.707, 0.311], [0.396, 0.749], [0.468, 0.579]]\\
		Customer demands: [5, 6, 3, 1, 9, 8, 9, 8, 7, 7]\\
		Depot location:  [0.204, 0.091]\\
		
		\midrule
		
		\textbf{Greedy decoder}:\\
		Tour Length: 5.420 \\
		Tour: 10 $\to$ 7 $\to$ 4 $\to$ 9 $\to$ 10 $\to$ 6 $\to$ 9 $\to$ 8 $\to$ 10 $\to$ 1 $\to$ 0 $\to$ 5 $\to$ 3 $\to$ 10 $\to$ 2 $\to$ 10\\
		\midrule
		\textbf{BS decoder with width 5}:\\
		Beam tour lengths: [5.697, 5.731, 5.420, 5.386, 5.582] \\
		Best beam: 3, Best tour length: 5.386\\
		Best tour: 10 $\to$ 7 $\to$ 4 $\to$ 6 $\to$ 10 $\to$ 6 $\to$ 8 $\to$ 9 $\to$ 10 $\to$ 1 $\to$ 0 $\to$ 5 $\to$ 3 $\to$ 10 $\to$ 2 $\to$ 10 \\
		\midrule
		
		\textbf{BS decoder with width 10}:\\
		Beam tour lengths: [5.697, 5.731, 5.420, 5.386, 5.362, 5.694, 5.582, 5.444, 5.333, 5.650 ]\\
		Best beam: 8 , Best tour length: 5.333\\
		Best tours: 10 $\to$ 7 $\to$ 4 $\to$ 9 $\to$ 10 $\to$ 9 $\to$ 6 $\to$ 8 $\to$ 10 $\to$ 1 $\to$ 0 $\to$ 5 $\to$ 3 $\to$ 10 $\to$ 2 $\to$ 10 \\
		\midrule
		\textbf{Optimal}: \\
		Optimal tour length: 6.037\\
		Optimal tour: 10 $\to$ 5 $\to$ 7 $\to$ 10 $\to$ 9 $\to$ 10 $\to$ 2 $\to$ 10 $\to$ 8 $\to$ 10 $\to$ 1 $\to$ 3 $\to$ 4 $\to$ 6 $\to$ 10\\
		\bottomrule
	\end{tabular}}
\end{table}
}

\subsection{Attention Mechanism Visualization}
\begin{figure*}[htbp]
	\centering
	\begin{subfigure}{.24\columnwidth}
		\centering
		\includegraphics[width=.8\columnwidth]{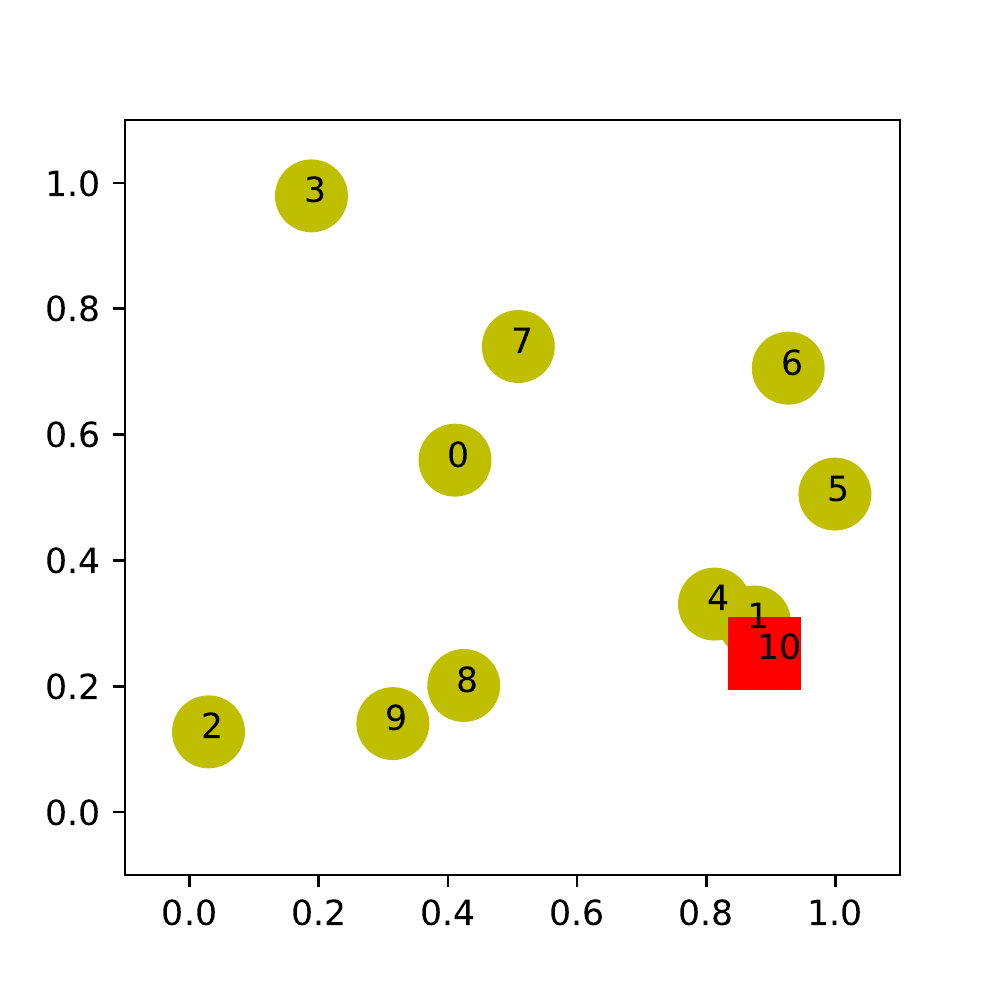}
		\caption{VRP10 instance.}
	\end{subfigure}
	\begin{subfigure}{.24\columnwidth}
		\centering
		\includegraphics[width=1.18\columnwidth]{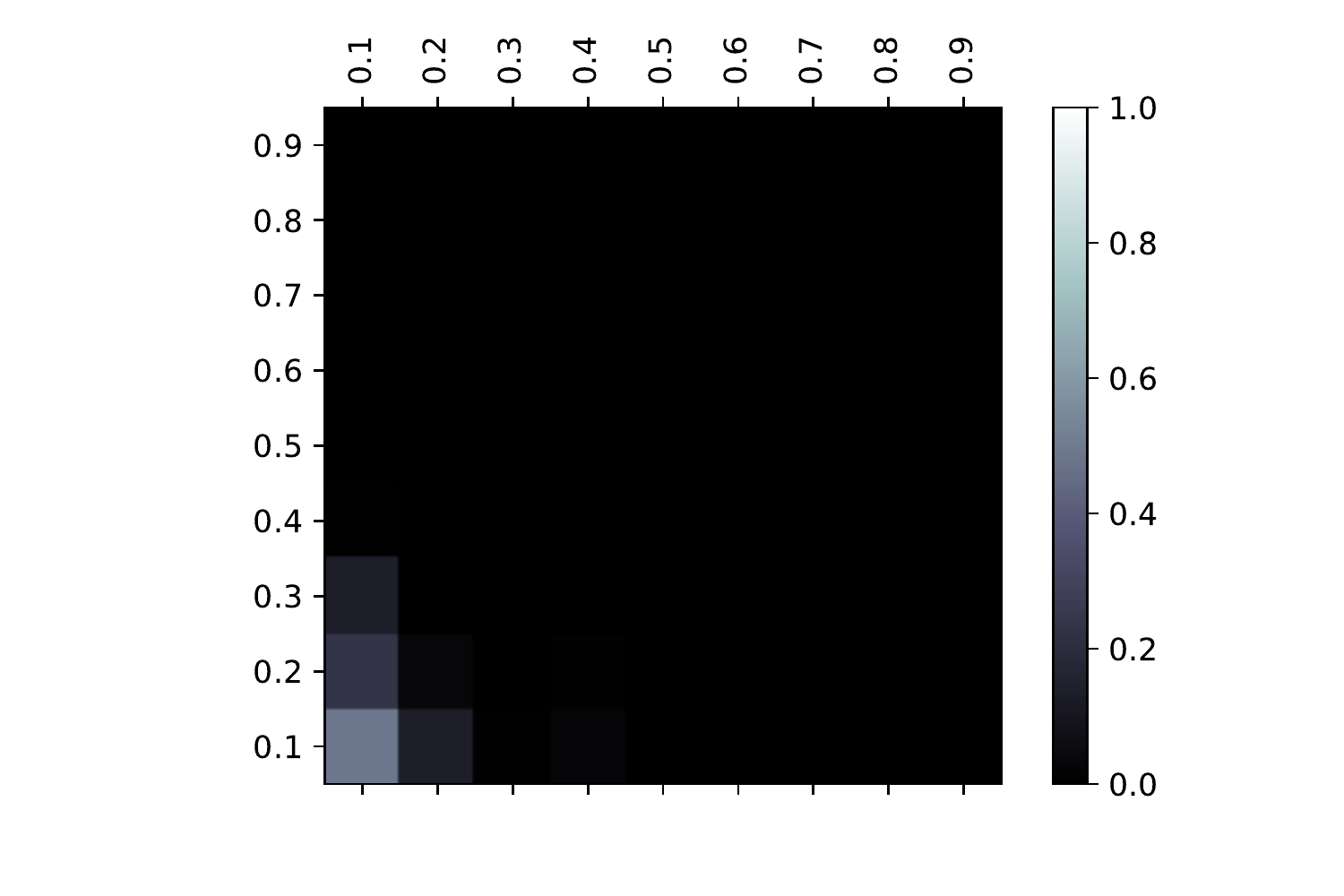}
		\caption{Point 0.}
	\end{subfigure}
	\begin{subfigure}{.24\columnwidth}
		\centering
		\includegraphics[width=1.18\columnwidth]{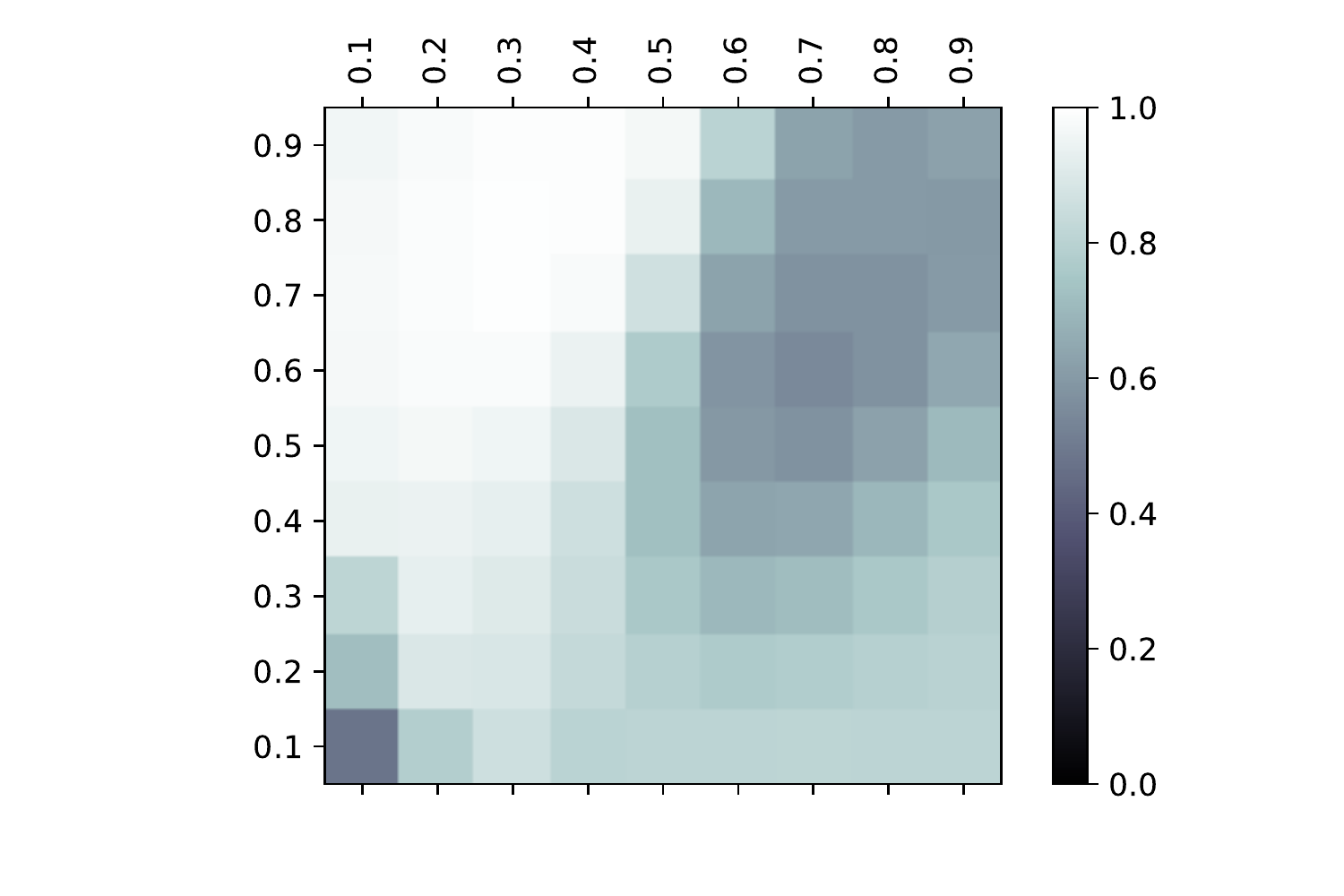}
		\caption{Point 2.}
	\end{subfigure} 
	\begin{subfigure}{.24\columnwidth}
		\centering
		\includegraphics[width=1.18\columnwidth]{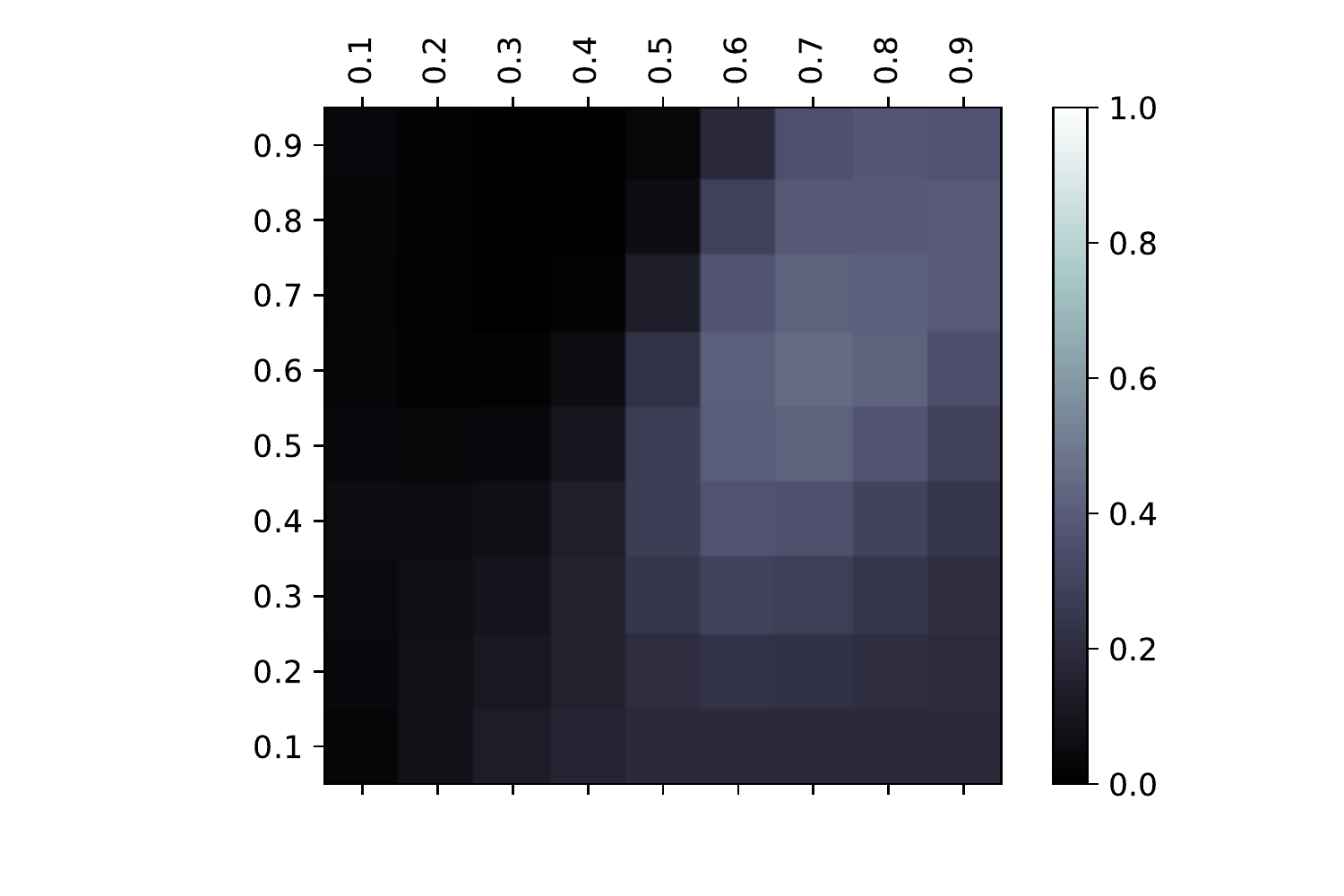}
		\caption{Point 9.}
	\end{subfigure} 
	\caption[Illustration of attention mechanism]{Illustration of attention mechanism at decoding step 0. The problem instance is illustrated in part (a) where the nodes are labeled with a sequential number; labels 0-9 are the customer nodes and 10 is the depot. We place node 0 at different locations and observe how it affects the probability distribution of choosing the first action, as illustrated in parts (b)--(d).}	
	\label{fig:vrp:att-illustration}
\end{figure*}

In order to illustrate how the attention mechanism is working, we relocated customer node 0 to different locations and observed how it affects the selected action. Figure \ref{fig:vrp:att-illustration} illustrates the attention in initial decoding step for a VRP10 instance drawn in part (a). Specifically, in this experiment, we let the coordinates of node 0 equal $\{0.1\times (i, j), \forall\, i,j\in \{1,\cdots,9\}\}$. In parts (b)-(d), the small bottom left square corresponds to the case where node 0 is located at [0.1,0.1] and the others have a similar interpretation. Each small square is associated with a color ranging from black to white, representing the probability of selecting the corresponding node at the initial decoding step. In part (b), we observe that if we relocate node 0 to the bottom-left of the plane, there is a positive probability of directly going to this node; otherwise, as seen in parts (c) and (d), either node 2 or 9 will be chosen with high probability. We do not display the probabilities of the other points since there is a near-0 probability of choosing them, irrespective of the location of node 0. A video demonstration of the model and attention mechanism is available online at \url{https://streamable.com/gadhf}.

\subsection{Experiment on Stochastic VRP}\label{sec:sto-vrp}

Next, we design a simulated experiment to illustrate the performance of the framework on the \textit{stochastic VRP} (SVRP). A major difficulty of planning in these systems is that the schedules are not defined in beforehand, and one needs to deal with various customer/demand realizations on the fly. Unlike the majority of the previous literature which only considers one stochastic element (e.g., customer locations are fixed, but the demands can change), we allow the customers and their demands to be stochastic, which makes the problem intractable for many classical algorithms. (See the review of SVRP by \citet{ritzinger2016survey}.) We consider an instance of the SVRP in which customers with random demands arrive at the system according to a Poisson process; without loss of generality we assume the process has rate 1. Similar to previous experiments, we choose each new customer's location uniformly on the unit square and its demand to a discrete number in $\{1,\cdots,9\}$. We fix the depot position to $[0.5,0.5]$. A vehicle is required to satisfy as much demand as possible in a time horizon with length 100 time units. To make the system stable, we assume that each customer cancels its demand after having gone unanswered for 5 time units. The vehicle moves with speed 0.1 per time unit. Obviously, this is a continuous-time system, but we view it as a discrete-time MDP where the vehicle can make decisions at either the times of customer arrivals or after the time when the vehicle reaches a node.

The network and its hyper-parameters in this experiment are the same as in the previous experiments. One major difference is the RL training method, where we use asynchronous advantage actor-critic (A3C) \cite{mnih2016asynchronous} with one-step reward accumulation. The main reason for choosing this training method is that REINFORCE is not an efficient algorithm in dealing with the long trajectories. The details of the training method are described in Appendix \ref{sec:vrp:app:a3c}. The other difference is that instead of using masking, at every time step, the input to the network is a set of available locations which consists of the customers with positive demand, the depot, and the vehicle's current location; the latter decision allows the vehicle to stop at its current position, if necessary. We also add the \textit{time-in-system} of customers as a dynamic element to the attention mechanism; it will allow the training process to learn customer abandonment behavior.

We compare our results with three other strategies: \textit{(i)} \textit{Random}, in which the next destination is randomly selected from the available nodes and it is providing a ``lower bound'' on the performance;
\textit{(ii)} \textit{Largest-Demand}, in which the customer with maximum demand will be chosen as the next destination; and \textit{(iii)} \textit{Max-Reachable}, in which the vehicle chooses the node with the highest demand while making sure that the demand will remain valid until the vehicle reaches the node. In all strategies, we force the vehicle to route to the depot and refill when its load is zero. Even though simple, these baselines are common in many applications. Implementing and comparing the results with more intricate SVRP baselines is an important future direction.

Table \ref{table:vrp:sto-vrp} summarizes the average demand satisfied, and the percentage of the total demand that this represents, under the various strategies, averaged over 100 test instances. We observe that A3C outperforms the other strategies. Even though A3C does not know any information about the problem structure, it is able to perform better than the Max-Reachable strategy, which uses customer abandonment information. 

\begin{table}[ht]
	\centering
	\caption{Satisfied demand under different strategies.}
	\label{table:vrp:sto-vrp}
	\begin{tabular}{m{1.6cm}m{1.15cm}m{1.2cm}m{1.2cm}m{1.05cm}}
		\toprule
		Method&	Random& Largest-Demand& Max-Reachable& A3C\\
		\midrule
		Avg. Dem. &24.83&75.11&88.60&112.21\\ %452.4
		\% satisfied &5.4\%&16.6\%&19.6\%&28.8\%
	\end{tabular}
\end{table}

\section{Training Policy Gradient Methods}
We utilize the REINFORCE method, similar to \citet{bello2016neural} for solving the TSP and VRP, and A3C \cite{mnih2016asynchronous} for the SVRP. In this Appendix, we explain the details of the algorithms.

Let us consider a family of problems, denoted by $\mathcal{M}$, and a probability distribution over them, denoted by $\Phi_\mathcal{M}$. During the training, the problem instances are generated according to distribution $\Phi_\mathcal{M}$. We also use the same distribution in the inference to produce test examples. 

\paragraph{REINFORCE Algorithm for VRP}
Algorithm \ref{vrp:alg-a2c} summarizes the REINFORCE algorithm. We have two neural networks with weight vectors $\theta$ and $\phi$ associated with the actor and critic, respectively. We draw $N$ sample problems from $\mathcal{M}$ and use Monte Carlo simulation to produce feasible sequences with respect to the current policy $\pi_\theta$. We adopt the superscript $n$ to refer to the variables of the $n$th instance. After termination of the decoding in all $N$ problems, we compute the corresponding rewards as well as the policy gradient in step 14 to update the actor network. In this step, $V(X_0^n;\phi)$ is the the reward approximation for instance problem $n$ that will be calculated from the critic network. We also update the critic network in step 15 in the direction of reducing the difference between the expected rewards with the observed ones during Monte Carlo roll-outs. 
\begin{algorithm}[htbp]
	\caption{REINFORCE Algorithm}
	\label{vrp:alg-a2c}
	
	\begin{algorithmic}[1]
		\STATE initialize the actor network with random weights $\theta$ and critic network with random weights $\phi$
		\FOR {$iteration = 1,2,\cdots$} 
		\STATE reset gradients: $d\theta\leftarrow 0$, $d\phi\leftarrow 0$
		\STATE sample $N$ instances according to $\Phi_\mathcal{M}$
		
		\FOR {$n=1,\cdots,N$}
		\STATE initialize step counter $t\leftarrow 0$
		\REPEAT
		\STATE choose $y_{t+1}^n$ according to the distribution $P(y_{t+1}^n| Y_{t}^n,{X}_{t}^n)$
		\STATE observe new state $X_{t+1}^n$
		\STATE $t \leftarrow t+1$
		\UNTIL{termination condition is satisfied}
		\STATE compute reward $R^n = R(Y^n,X^n_0) $
		\ENDFOR
		\STATE $d\theta \leftarrow \frac{1}{N} \sum_{n=1}^N \left(R^n - V(X_0^n;\phi)\right) \nabla_\theta \log P(Y^n|X_0^n)$\label{eq:vrp:alg:policy_update}
		\STATE $d\phi \leftarrow \frac{1}{N} \sum_{n=1}^N \nabla_\phi \left(R^n - V(X_0^n;\phi)\right)^2$\label{eq:vrp:alg:value}

		\STATE update $\theta$ using $d\theta$ and $\phi$ using $d\phi$.
		\ENDFOR
	\end{algorithmic}
\end{algorithm}

\paragraph{Asynchronous Advantage Actor-Critic for SVRP}\label{sec:vrp:app:a3c}
The \textit{Asynchronous Advantage Actor-Critic} (A3C) method proposed in \cite{mnih2016asynchronous} is a policy gradient approach that has been shown to achieve super-human performance playing Atari games. In this paper, we utilize this algorithm for training the policy in the SVRP. In this architecture, we have a central network with weights $\theta^0,\phi^0$ associated with the actor and critic, respectively. In addition, $N$ agents are running in parallel threads, each having their own set of local network parameters; we denote by $\theta^n,\phi^n$ the actor and critic weights of thread $n$. (We will use superscript $n$ to denote the operations running on thread $n$.) Each agent interacts with its own copy of the VRP at the same time as the other agents are interacting with theirs; at each time-step, the vehicle chooses the next point to visit and receives some reward (or cost) and then goes to the next time-step. In the SVRP that we consider in this paper, $R_t$ is the number of demands satisfied at time $t$. We note that the system is basically a continuous-time MDP, but in this algorithm, we consider it as a discrete-time MDP running on the times of system state changes $\{\tau_t: t=0,\cdots\}$; for this reason, we normalize the reward $R_t$ with the duration from the previous time step, e.g., the reward is $R_t/(\tau_t-\tau_{t-1})$. The goal of each agent is to gather independent experiences from the other agents and send the gradient updates to the central network located in the main thread. In this approach, we periodically update the central network weights by accumulated gradients and send the updated weight to all threads. This asynchronous update procedure leads to a smooth training since the gradients are calculated from independent VRP instances. 

\begin{algorithm}[tb]
	\caption{Asynchronous Advantage Actor-Critic (A3C)}
	\label{vrp:alg-a3c}
	
	\begin{algorithmic}[1]
		
		\STATE initialize the actor network with random weights $\theta^0$ and critic network with random weights $\phi^0$ in the master thread.
		\STATE initialize $N$ thread-specific actor and critic networks with weights $\theta^n$ and $\phi^n$ associated with thread $n$.
		\REPEAT
		\FOR {each thread $n$ }

		\STATE sample a instance problem from $\Phi_\mathcal{M}$ with initial state $X_0^n$
		\STATE initialize step counter $t^n\leftarrow 0$
		\WHILE{episode not finished}
		
		\STATE choose $y_{t+1}^n$ according to $P(y_{t+1}^n| Y_{t}^n,{X}_{t}^n;\theta^n)$
		\STATE observe new state $X_{t+1}^n$; 
		\STATE observe one-step reward $R^n_t = R(Y^n_{t},X^n_{t})$
		\STATE let $A_t^n =\left(R^n_t+ V(X_{t+1}^n;\phi) - V(X_{t}^n;\phi)\right)$
		\STATE $d\theta^0 \leftarrow d\theta^0+   \nabla_\theta A_t^n\log P(y_{t+1}^n| Y_{t}^n,{X}_{t}^n;\theta^n)$
		\STATE $d\phi^0 \leftarrow d\phi^0 +\nabla_\phi \left(A_t^n\right)^2$
		\STATE $t^n \leftarrow t^n+1$
		\ENDWHILE

		\ENDFOR
		\STATE periodically update $\theta^0$ using $d\theta^0$ and $\phi^0$ using $d\phi^0$
		\STATE  $\theta^n  \leftarrow \theta^0$, $\phi^n  \leftarrow \phi^0$
		\STATE reset gradients: $d\theta^0\leftarrow 0$, $d\phi^0\leftarrow 0$
		
		\UNTIL{training is finished}
		
	\end{algorithmic}
\end{algorithm}

Both actor and critic networks in this experiment are exactly the same as the ones that we employed for the classical VRP. For training the central network, we use RMSProp optimizer with learning rate $10^{-5}$.

\end{document}